\pdfoutput=1

\documentclass[11pt]{article}

\usepackage[]{acl}
\usepackage{multirow}
\usepackage{graphicx}
\usepackage{booktabs}
\usepackage{tabularx}
\usepackage{CJKutf8}
\usepackage{amssymb}
\usepackage{amsmath}
\usepackage{tabu} 
\usepackage{hyperref}
\usepackage{breakurl}
\usepackage{times}
\usepackage{latexsym}
\usepackage{tikz}
\usepackage{pgfplots}
\usepackage[normalem]{ulem}
\usepackage{xcolor}
\usepackage{dashrule}
\usepackage{colortbl}
\usepackage{tipa}

\definecolor{brickred}{HTML}{b92622}
\definecolor{midnightblue}{HTML}{005c7f}
\definecolor{salmon}{HTML}{f1958d}
\definecolor{burntorange}{HTML}{f19249}
\definecolor{junglegreen}{HTML}{4dae9d}
\definecolor{forestgreen}{HTML}{499c5e}
\definecolor{pinegreen}{HTML}{3d8a75}
\definecolor{seagreen}{HTML}{6bc1a2}
\definecolor{limegreen}{HTML}{97c65a}
\definecolor{redtable}{HTML}{E67F72}
\definecolor{orangetable}{HTML}{E67F72}
\definecolor{greentable}{HTML}{E67F72}

\usepackage{lipsum}

\newcommand\blfootnote[1]{%
  \begingroup
  \renewcommand\thefootnote{}\footnote{#1}%
  \addtocounter{footnote}{-1}%
  \endgroup
}

\usepackage[T1]{fontenc}

\usepackage[utf8]{inputenc}

\usepackage{microtype}

\title{NaSGEC: a Multi-Domain Chinese Grammatical Error Correction Dataset \\ from Native Speaker Texts}

\author{Yue Zhang$^{1}$, Bo Zhang$^{2}$, Haochen Jiang$^1$, Zhenghua Li$^{1*}$, {\bf Chen Li$^2$}, {\bf Fei Huang$^2$}, {\bf Min Zhang$^1$} \\
        $^1$Institute of Artificial Intelligence, School of Computer Science and Technology, \\
Soochow University, China; $^2$DAMO Academy, Alibaba Group, China\\
\texttt{$^1$\{yzhang21,hcj22\}@stu.suda.edu.cn}, \texttt{$^1$\{zhli13,minzhang\}@suda.edu.cn}\\\texttt{$^2$\{klayzhang.zb,puji.lc,f.huang\}@alibaba-inc.com}}

\begin{document}
\begin{CJK}{UTF8}{gkai}

\maketitle


\begin{abstract}
We introduce NaSGEC, a new dataset to facilitate research on Chinese grammatical error correction (CGEC) for native speaker texts from multiple domains. Previous CGEC research primarily focuses on correcting texts from a single domain, especially learner essays. To broaden the target domain, we annotate multiple references for 12,500 sentences from three native domains, i.e., social media, scientific writing, and examination. We provide solid benchmark results for NaSGEC by employing cutting-edge CGEC models and different training data. We further perform detailed analyses of the connections and gaps between our domains from both empirical and statistical views. We hope this work can inspire future studies on an important but under-explored direction---cross-domain GEC.

\end{abstract}
\section{Introduction}

Grammatical error correction (GEC) aims to remove all underlying textual errors in a given sentence without changing its meaning \cite{bryant2022survey}. During the past decade, GEC has attracted a lot of research interest and has been integrated into many real-life applications like writing assistants.\blfootnote{$^*$ Corresponding author.}


A significant effort has been undertaken to build high-quality datasets for research on GEC. Most GEC datasets are for English \cite{yannakoudakis2011new, dahlmeier2013building, napoles2017jfleg,bryant2019bea}, which mainly collect sentences from learner essays. For Chinese GEC (CGEC), datasets are relatively scarce. Similar to English GEC, most of them are built from essays written by learners, including NLPCC18 \cite{zhao2018overview}, CGED \cite{rao2018overview, rao2020overview}, YACLC \cite{wang2021yaclc}, and MuCGEC \cite{DBLP:conf/naacl/0004LBLZLHZ22}. 

Besides learner GEC, there is also great demand for correcting errors made by native speakers. For English GEC, researchers have already constructed several native datasets, e.g., GMEG \cite{napoles2019enabling} and CWEB \cite{flachs2020grammatical}. For CGEC, such research has just begun. CCTC \cite{wang2022cctc} is the first native CGEC dataset composed of web documents written by natives. Another recent work, FCGEC \cite{xu2022FCGEC}, collects sentences from the questions in Chinese examinations.


\begin{table}[]
\centering
\scalebox{0.9}{
\begin{tabular}{ll}
\toprule
\textbf{Source} & \begin{tabular}[c]{@{}l@{}}目前现有的汉语依存书库规模较小。\\ The scale of current existing Chinese \\ dependency library is relatively small. \end{tabular} \\ \hline
\textbf{Ref. 1} & \begin{tabular}[c]{@{}l@{}}\sout{目前}现有的汉语依存\textcolor{red}{树}库规模较小。\\ The scale of \sout{current} existing Chinese \\ dependency \textcolor{red}{treebank} is relatively small. \end{tabular}        \\ \hline
\textbf{Ref. 2} & \begin{tabular}[c]{@{}l@{}}目前\sout{现有}的汉语依存\textcolor{red}{树}库规模较小。\\ The scale of current \sout{existing} Chinese  \\ dependency \textcolor{red}{treebank} is relatively small. \end{tabular}       \\
\bottomrule 
\end{tabular}
}

\caption{A native CGEC example with two references from the \textsc{Thesis} domain of NaSGEC.}
\label{tab:example:1}
\end{table}

Among all the above datasets, only GMEG \cite{napoles2019enabling} targets texts from multiple domains.
The lack of multi-domain datasets inevitably introduces biases in the construction and evaluation of CGEC approaches. 
First, cutting-edge CGEC approaches  \cite{DBLP:conf/aaai/LiGZSJRX22, zhang2022syngec, wu2022from} are all evaluated under the in-domain setting, where the training and test sets are from the same domain. 
It remains unclear how well those approaches generalize to out-of-domain inputs, which is important for practical application.
Second, all CGEC approaches are only evaluated in a single domain, basically learner essays. This can be misleading since an approach that outperforms others in one domain may actually perform poorly in another.

To alleviate these problems, this work proposes \textbf{NaSGEC} (pronounced as \textipa{/"neIsgek/}), a multi-domain dataset from \textbf{na}tive \textbf{s}peaker texts for Chinese \textbf{GEC}. NaSGEC comprises 12,500 sentences from 3 native text domains: social media platform (\textsc{Media}), undergraduate theses (\textsc{Thesis}), and Chinese examinations (\textsc{Exam}).
These domains are closely related to real-life GEC application scenarios, i.e., writing aid, paper proofreading, and Chinese teaching.
Based on detailed data analysis (see Section \ref{sec:ana}), we demonstrate that they have diverse writing styles and error distributions, thus posing great challenges for existing models and will be an ideal testbed for domain adaptation techniques.
Furthermore, there are usually different correction methods for an error, as shown in Table \ref{tab:example:1}.
Hence, we assign each sentence to two annotators for annotation and one expert for double-checking, leading to multiple high-quality references.







\begin{figure}[tp!]
\centering
\includegraphics[scale=0.4]{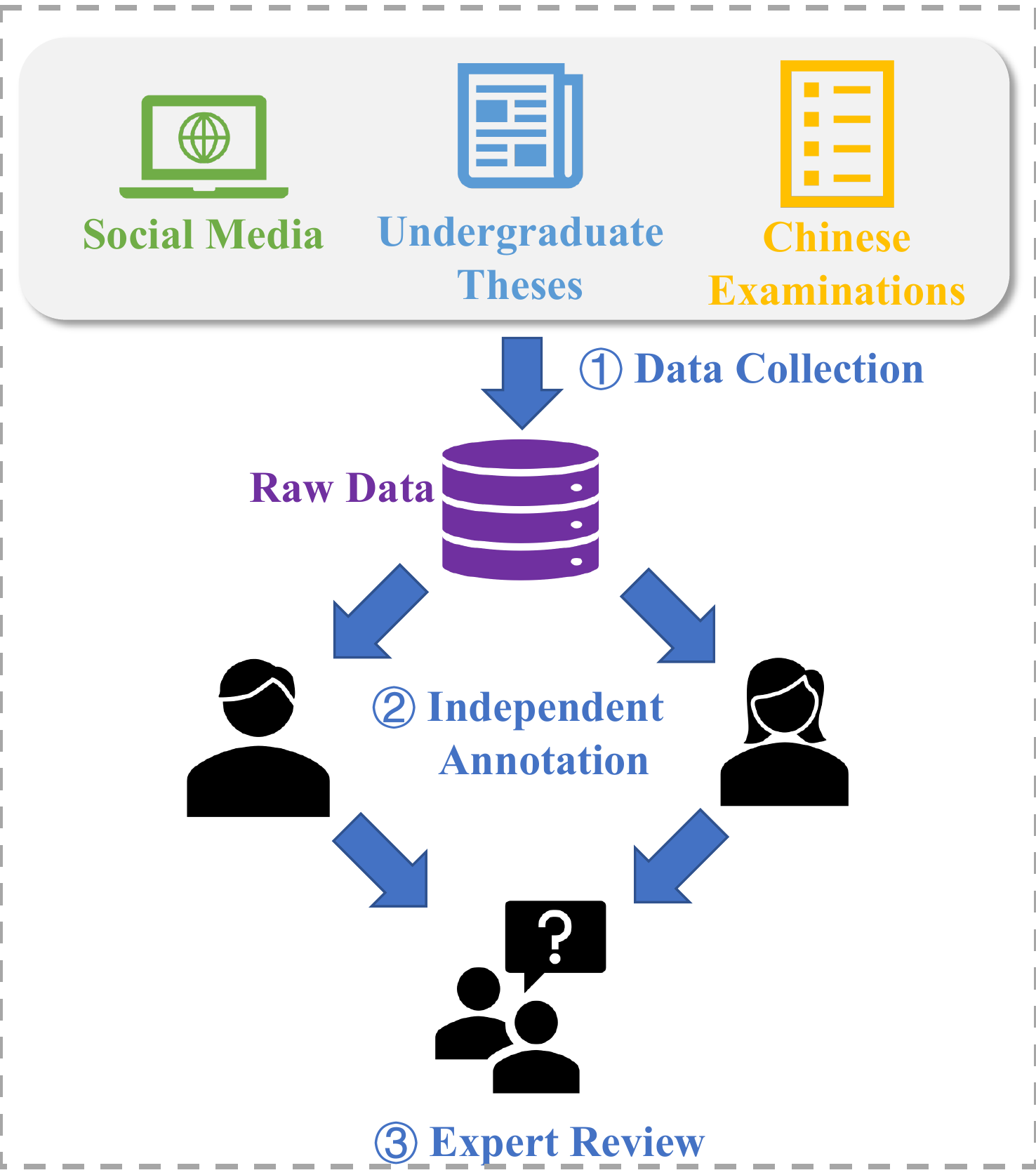}
\caption{The construction procedure of NaSGEC.}
\label{fig:anno:flow}
\end{figure}
Using NaSGEC, we conduct extensive experiments. 
We evaluate the performance of the state-of-the-art (SOTA) CGEC model on NaSGEC with different kinds of training data.
We first train the model on commonly-used human-annotated training sets. 
Since these training sets are collected from learner texts while NaSGEC is a native dataset, we also generate synthetic training data from native texts.
The multi-domain property of NaSGEC enables us to shed light on the domain problem in CGEC.
We conduct domain transfer experiments and design three indicators for evaluating domain differences. 
In summary, our main contributions can be concluded as follows:

\begin{itemize}
    \item[(1)] We propose NaSGEC, a multi-domain CGEC dataset from native speaker texts, which contains 12.5k sentences with multiple references. We also conduct detailed data analysis on it.

    \item[(2)] We launch benchmark experiments on NaSGEC with 
    SOTA CGEC models and different training data. 
    We find models have their own advantages in specific domains, suggesting that the multi-domain NaSGEC can support a more comprehensive evaluation.

    \item[(3)] Based on NaSGEC, we perform preliminary domain transfer experiments and analysis. We find using small-scale in-domain data for fine-tuning can significantly boost model performance. We also analyze the similarity between domains by comparing cross-domain transfer performance. We devise several indicators of domain shifts to gain more insights. To further improve model performance in a specific domain, we propose a simple domain-aware data augmentation method.

    \item[(4)] We systematically compare NaSGEC to previously released CGEC datasets, including both learner and native ones. 
    
\end{itemize}

All codes and models have been released at \url{https://github.com/HillZhang1999/NaSGEC}. We will also release the dataset after improving it according to reviewers' comments. 

\begin{table*}[!htbp]
\begin{center}
\centering
\scalebox{0.7}{
\begin{tabular}{llcccccccc}
\toprule
\textbf{Dataset} & \textbf{Writer} & \textbf{\#Sent.} & \textbf{\#Err. Sent. (Perc.)} & \textbf{Avg. Length} & \textbf{Avg. Edits} & \textbf{Avg. Refs} & \textbf{Avg. NEs} & \textbf{Type-Token}   \\ \hline
\textbf{NLPCC18 \cite{zhao2018overview}}   & Learner      & 2,000              & 1,983 (99.2\%)           & 29.7              & 2.0                & 1.1 & 0.39  & 0.43                             \\
\textbf{MuCGEC \cite{DBLP:conf/naacl/0004LBLZLHZ22}}  & Learner      & 7,063              & 6,544 (92.7\%)           & 38.5              & 3.2                & 2.3 & 0.38  & 0.42                             \\ \hline
\textbf{CCTC \cite{wang2022cctc}}   & Native     & 25,207              & 2,331 (9.3\%)           & 41.8               & 1.0                & 1.0  & 0.68  & 0.53                          \\
\textbf{FCGEC \cite{xu2022FCGEC}}   & Native     & 41,340              & 22,517 (54.6\%)           & 53.1               & 1.5                & 1.7  & 1.91  & 0.49                           \\
\hline\hline
\textbf{NaSGEC (\textsc{Media})}  & Native    & 4,000             & 2,605 (65.2\%)            & 49.0               & 1.8                & 1.4  & 0.79  & 0.55                       \\
\textbf{NaSGEC (\textsc{Thesis})}  & Native      &  1,500            & 1,050 (70.0\%)           & 60.5               & 1.9                & 1.5  & 0.67  & 0.45                          \\
\textbf{NaSGEC (\textsc{Exam})} & Native      & 7,000             & 4,849 (69.3\%)           & 55.9               & 1.4                & 1.7  & 1.00  & 0.51                           \\ \hline
\textbf{NaSGEC}   & Native     & 12,500              & 8,504 (68.0\%)           & 54.3               & 1.6                & 1.6    & 0.89  & 0.52                     \\ \bottomrule
\end{tabular}
}
\caption{Dataset statistics, including the writer, the number of sentences (\#Sent.), the number and percentage of erroneous sentences (\#Err. Sent. (Perc.)), the average length (characters) of sentences (Avg. Length), the average number of edits per reference (Avg. Edits), the average number of references (Avg. Refs), the average number of named entities per sentence (Avg. NEs, extracted by the LTP 
toolkit \cite{DBLP:conf/coling/CheLL10}), the average ratio of vocabulary size by the total number of
tokens (Type-token, calculated following \citet{flachs2020grammatical}). 
}
\label{tab:overall:statistic}
\end{center}
\end{table*}
\section{Construction of NaSGEC}

This section describes the construction process of NaSGEC in detail. As shown in Figure \ref{fig:anno:flow}, we first collect raw sentences from three domains. Then, each sentence is assigned to two annotators for independent annotation. To guarantee data quality, an expert will carefully review the annotation results.

\subsection{Data Collection}
\label{sec:dc}
NaSGEC collects data from 3 native Chinese text domains, which cover both formal and informal writing styles and errors of different difficulties.

The \textbf{\textsc{Media}} domain contains 4k sentences from articles posted on the \textit{Wechat public account platform}\footnote{\url{https://mp.weixin.qq.com/}}, which is one of the most popular social media platforms in China. 
Articles in this platform covers rich topics. 
We also notice that the sentences in it are mostly informal and often expressed in a spoken-language tone.
During our preliminary annotation, we found that errors in this domain are extremely sparse, so direct annotation would result in high costs to acquire enough erroneous sentences. 
Therefore, we turn to select sentences by voting with multiple competitive CGEC models.
Specifically, we utilize large-scale pseudo training data to train three seq2seq-based models and three seq2edit-based models.
Then, we only choose candidate sentences corrected by more than half of those models for annotation. We crawl 1M candidate sentences from the Wechat public account platform, and accumulate about 120k potentially wrong sentences from them with the above-mentioned method. Finally, we randomly pick 4k sentences for annotation.

The \textbf{\textsc{Thesis}} domain consists of 1.5k sentences from \textit{undergraduate theses}. We first collect 120 dissertations written by Chinese undergraduates majoring in computer science, with about 40k sentences in total. Intuitively, texts in this domain are usually formal and contain technical terms. Similar to \textsc{Media}, errors in \textsc{Thesis} are also very sparse. To save costs, we adopt the same method as in \textsc{Media} to select sentences for annotation.

The \textbf{\textsc{Exam}} domain contains 7k sentences from the \textit{ungrammatical sentence judgment questions in Chinese examinations.} Such questions are elaborately designed by experts and ask students to choose 1-3 ungrammatical sentences from 4 candidates. We crawl them from a public educational website\footnote{\url{http://www.gzywtk.com/}}, as well as their answers and analyses. 

\subsection{Annotation Workflow}
\label{sec:anno:flow}
For groundwork, we extend the annotation guidelines of MuCGEC \cite{DBLP:conf/naacl/0004LBLZLHZ22} 
to accommodate errors made by native speakers. 
We subsequently use them to instruct our annotators and gradually improve them according to annotator feedback before starting the annotation process.
For example, we define how to distinguish dialect from errors after discussing with annotators.

During annotation, we ask our annotators to directly rewrite the whole sentence to craft a grammatical and fluent version of it with its intended meaning. The so-called \textit{direct rewriting} annotation paradigm has proven efficient and effective in GEC \cite{sakaguchi2016reassessing, napoles2017jfleg}.

Since multiple acceptable correction ways usually exist, we assign each sentence to two random annotators for independent annotation.
Following 
\citet{DBLP:conf/naacl/0004LBLZLHZ22}, 
we ask each annotator to submit the best reference in his/her mind to improve the annotation efficiency.
Then, 
an expert reviewer will check these two submissions in a double-blind manner. 
Besides directly rejecting incorrect submissions, the reviewer also needs to supplement other correct references missed by annotators. 
If annotators make wrong submissions, they are required to learn from their mistakes for self-improvement.
The learning method is re-typing one of the correct references determined by reviewers. 
All annotations are conducted with the support of our developed online annotation platform, which is presented in Appendix \ref{sec:anno:tool}. We select and show some typical annotation examples in Appendix \ref{sec:examples}.



\definecolor{Color1}{HTML}{71AC47}
\definecolor{Color2}{HTML}{5B9AD4}
\definecolor{Color3}{HTML}{FBBF02}
\definecolor{Color4}{HTML}{022061}

\begin{figure*}[t!]
\begin{tikzpicture}
  \centering
  \begin{axis} [    
    ybar, 
    axis on top, 
    bar width=0.22cm,
    ymajorgrids, tick align=inside,
    major y grid style={densely dashed},
    width=15cm,
    height=6.5cm,
    ymin=0, ymax=70,
    tickwidth=0pt,
    legend style={
        draw=none,
        at={(.43,0.99)},
        legend columns=-1,
        anchor=north,
        /tikz/every even column/.append style={column sep=.3cm}
    },
    ylabel={Percentage (\%)},
    ytick={0, 10, 20, ..., 70},
        symbolic x coords={
      \small{NaSGEC$_{\textsc{Media}}$}, \small{NaSGEC$_{\textsc{Thesis}}$}, \small{NaSGEC$_{\textsc{Exam}}$}, \small{MuCGEC}, \small{FCGEC}, \small{CCTC}
    },
    xtick=data,
    legend image code/.code={
        \draw [#1] (-0.15cm,-0.15cm) rectangle (0.20cm,0.20cm); 
    },
  ]
  
  \addplot [draw=Color1, fill=Color1] coordinates {
    (\small{NaSGEC$_{\textsc{Media}}$},  47.94)
    (\small{NaSGEC$_{\textsc{Thesis}}$},  50.44)
    (\small{NaSGEC$_{\textsc{Exam}}$},  26.56)
    (\small{MuCGEC},   37.92)
    (\small{CCTC},   67.95)
    (\small{FCGEC},   21.27)
  };

  \addplot [draw=Color2, fill=Color2] coordinates {
    (\small{NaSGEC$_{\textsc{Media}}$},  35.26)
    (\small{NaSGEC$_{\textsc{Thesis}}$},  25.60)
    (\small{NaSGEC$_{\textsc{Exam}}$},  30.59)
    (\small{MuCGEC},   36.43)
    (\small{CCTC},   15.45)
    (\small{FCGEC},   27.11)
  };

  \addplot [draw=Color3, fill=Color3]
 coordinates {
    (\small{NaSGEC$_{\textsc{Media}}$},  14.14)
    (\small{NaSGEC$_{\textsc{Thesis}}$},  20.69)
    (\small{NaSGEC$_{\textsc{Exam}}$},  30.50)
    (\small{MuCGEC},   20.06)
    (\small{CCTC},   14.98)
    (\small{FCGEC},   36.80)
  };

  \addplot [draw=Color4, fill=Color4]
  coordinates {
    (\small{NaSGEC$_{\textsc{Media}}$},  2.67)
    (\small{NaSGEC$_{\textsc{Thesis}}$},  3.27)
    (\small{NaSGEC$_{\textsc{Exam}}$},  12.35)
    (\small{MuCGEC},   5.59)
    (\small{CCTC},   4.71)
    (\small{FCGEC},   14.82)
  };

  \legend{\small{Substituted Error}, \small{Missing Error}, \small{Redundant Error}, \small{Word-order Error}}
  \end{axis}
\end{tikzpicture}
    \caption{The distributions of 4 kinds of error in 3 domains of NaSGEC and other CGEC datasets.}
\label{fig:error:dis}
\end{figure*}
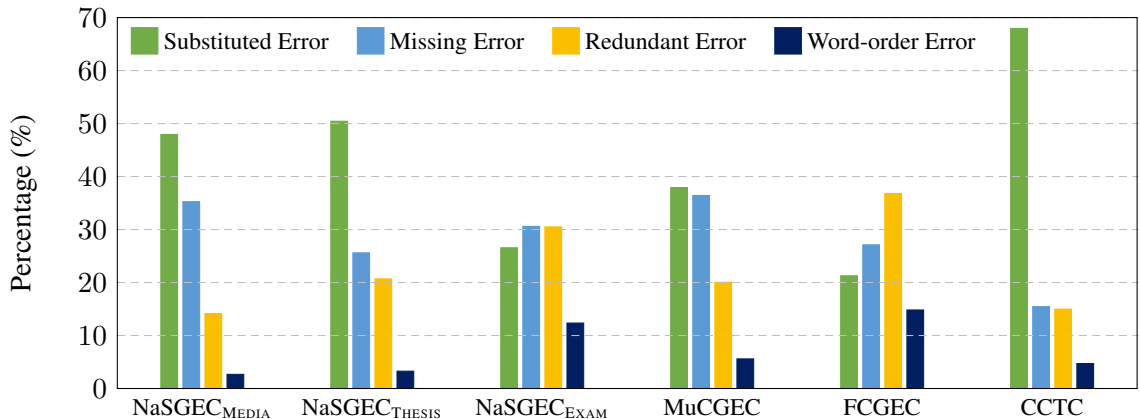


\subsection{Annotation Process}
We hired 13 well-educated native undergraduates familiar with Chinese grammar as our annotators. 2 graduate students, who participated in the compilation of guidelines, served as the reviewers. Annotators received detailed instructions before annotating; those with low annotation quality were warned during annotating. We established a chat group to allow annotators to ask questions. All annotators and reviewers were paid properly. The whole annotation process took more than 4 months.

\section{Analysis of NaSGEC}
\label{sec:ana}


\paragraph{Overall statistics.} We list detailed statistics of NaSGEC and other existing datasets for comparison in Table \ref{tab:overall:statistic}.
We use the tool\footnote{\url{https://github.com/HillZhang1999/MuCGEC/tree/main/scorers/ChERRANT}} released with MuCGEC \cite{DBLP:conf/naacl/0004LBLZLHZ22} to extract the edits of references and original sentences.
Such edits are span-level edits merged from character-based ones based on pre-defined linguistic rules.


Within NaSGEC, the average length of sentences varies across domains. The sentences in \textsc{Thesis} are the longest, probably because students tend to write long sentences in dissertations to explain technical concepts more clearly. 
Regarding the average number of edits and references, we observe that erroneous sentences in \textsc{Exam} need the fewest edits to correct but have the most correction ways.
The reason may be that each erroneous sentence in \textsc{Exam} typically only has one complicated error to challenge students, which is often varied in its valid corrections. 
As reflected by the type-token ratio \cite{richards1987type}, \textsc{Media} has the greatest lexical variety, intuitively due to the diversity of its topics. All the above analysis indicates systematical discrepancies across NaSGEC's domains.

We also present the statistics of two mainstream learner datasets, i.e., NLPCC18 \cite{zhao2018overview} and MuCGEC \cite{DBLP:conf/naacl/0004LBLZLHZ22}. Compared with those learner datasets, sentences in NaSGEC are significantly longer but contain much fewer edits, as natives make mistakes far less frequently than learners and seldom make obvious mistakes. Besides, sentences in NaSGEC also have more name entities and a higher level of lexical variety, showing that natives have a larger vocabulary.

Moreover, we also compare two newly published native datasets, CCTC \cite{wang2022cctc} and FCGEC \cite{xu2022FCGEC}.
The salient feature of CCTC is its low error density. Only 9.3\% of sentences in CCTC contain errors, and each erroneous sentence just has one error (reflected by Avg. Edits). 
As for FCGEC, it is quite similar to the \textsc{Exam} domain of NaSGEC, which is unsurprising since they share the same provenance. 


\textbf{
\begin{table*}[t!]
\centering
\scalebox{0.73}{
\begin{tabular}{l|ccc|ccc|ccc|ccc}
\toprule
 & \multicolumn{3}{c|}{\textbf{\textsc{Media}}}    & \multicolumn{3}{c|}{\textbf{\textsc{Thesis}}}   & \multicolumn{3}{c|}{\textbf{\textsc{Exam}}}   & \multicolumn{3}{c}{\textbf{Average}} \\ 
& \textbf{P} & \textbf{R} & \textbf{F$_{0.5}$} & \textbf{P} & \textbf{R} & \textbf{F$_{0.5}$} & \textbf{P} & \textbf{R} & \textbf{F$_{0.5}$}  & \textbf{P} & \textbf{R} & \textbf{F$_{0.5}$} \\ \midrule
\textbf{Real Learner}                                                                        & 35.96 & 29.15 & 34.35 & 24.16 & 34.06 & 25.65 & 23.01 & 11.31 & 19.06  & 27.71 & 24.84 & 27.08  \\
\textbf{Pseudo Native}                                                                       & \textbf{53.39} & 29.17 & \textbf{45.79} & 30.86 & 33.52 & 31.15 & 9.78 & 2.60 & 6.30  & 31.34 & 21.76 & \textbf{28.80} \\
\textbf{Pseudo Native $\Rightarrow$ Real Learner} & 38.37 & \textbf{31.16} & 36.67 & 25.67 & \textbf{35.09} & 27.13 & \textbf{24.48} & \textbf{11.59} & \textbf{20.02}  & 29.51 & \textbf{25.95} & 28.72 \\
\textbf{Real Learner $\Rightarrow$ Pseudo Native} & 51.90 & 26.20 & 43.39 & \textbf{31.61} & 31.97 & \textbf{31.87} & 10.77 & 2.52 & 6.51  & \textbf{31.43} & 20.23 & 28.29 \\ \bottomrule
\end{tabular}
}
\caption{Benchmark results on NaSGEC. ``Pseudo Native $\Rightarrow$ Real Learner'' means that we first train the model on pseudo native data, then on real learner data. The same goes for ``Real Learner $\Rightarrow$ Pseudo Native''.}
\label{tab:data:exp}
\end{table*}}
\vspace{-0.5cm}

\paragraph{Error type distributions.} 
We use the tool provided by MuCGEC to classify extracted edits into 4 error types according to their correction operations.
Figure \ref{fig:error:dis} shows the distributions of these error types in NaSGEC and other datasets for comparison.

Within NaSGEC, the most frequent error type in \textsc{Media} and \textsc{Theis} is substituted errors. After further decomposition, we find that the majority of substituted errors in these 2 domains are caused by spelling or misuse of punctuation, as native speakers usually make such minor mistakes due to carelessness when typing essays or papers.
The \textsc{Media} domain also has a considerable proportion of missing errors, mainly caused by missing punctuation. Such errors often occur in informal texts, as the absence of punctuation generally does not affect the understanding of the sentence. Compared with the other domains, \textsc{Exam} has a more even type distribution, where the proportion of substituted, missing, and redundant errors is quite close.

Like \textsc{Media} and \textsc{Thesis} domains of NaSGEC, the learner dataset MuCGEC also has a high proportion of substituted and missing errors. After a deeper look into samples, we find that learners are more prone to misuse verbs or nouns due to lexical or grammatical unfamiliarity, and they also tend to miss more specific words instead of punctuation.

Among all datasets, CCTC has the most unbalanced distribution: the substituted errors account for nearly 70\%, and we find most of them are caused by spelling. Although both come from Chinese examinations, FCGEC and NaSGEC-\textsc{Exam} still have some discrepancies, such as FCGEC contains more redundant errors, which may be due to different annotation guidelines and data sources.

\paragraph{Annotation Accuracy.} We measure each annotator's accuracy by comparing all his/her submissions against the golden references determined by reviewers. Overall, the average annotation accuracy is 77.46\%. Such a low figure clearly indicates the difficulty of the CGEC task. Moreover, it also highlights the importance of our review mechanism: about a quarter of references in our dataset will be problematic without our strict expert checking.

\section{Benchmark Experiments on NaSGEC}
\label{sec:ben}
This section provides benchmark results for NaSGEC with a current SOTA CGEC model. 
Following previous work, we train the model on human-annotated training data from learner texts. 
However, there exists a gap between learner training data and our native dataset. 
So we also use synthetic native training data to mitigate the gap.




\subsection{Experimental Setup}

\paragraph{Model.} 
Our benchmark models are based on BART \cite{lewis2020bart}, a pre-trained Seq2Seq model that has recently achieved SOTA performance on mainstream CGEC datasets \cite{zhang2022syngec, wu2022from}\footnote{We also experiment with another competitive CGEC paradigm (Seq2Edit) and report results in Appendix \ref{sec:seq2edit}.}.
We provide the implementation and training details in Appendix \ref{sec:exp:detail}. 



\paragraph{Evaluation metric.} We use the character-based metric proposed by \citet{DBLP:conf/naacl/0004LBLZLHZ22}.
Concretely, we align the system output and golden reference with the input sentence to extract two groups of character-based edits.
Then, we merge them into spans based on rules and compare them to calculate the precision (P), recall (R), and F$_{0.5}$ score.
In the GEC community, there is a consensus that a good system should correct errors accurately to ensure a positive user experience. Therefore, most work uses F$_{0.5}$, which places more emphasis on precision by weighting precision twice as recall.
We do not use previous word-based metrics since we find they will introduce uncertainty into evaluation due to word segmentation errors. 



\subsection{Training Data}
\label{sec:train:data}

\paragraph{Real learner training data.} There are two public available large-scale human-annotated CGEC training datasets, which refer to HSK \cite{zhang2009hsk} and Lang8 \cite{zhao2018overview}. 
Both of them focus on errors occurring in learner essays. 
Lang8 has about 1.2M sentence pairs, and HSK contains about 150k. 
We combine them together for training and randomly select 5k of them as the dev set following previous work \cite{DBLP:conf/naacl/0004LBLZLHZ22}.

\paragraph{Pseudo native training data.} So far, there has been no large-scale training data for errors made by native speakers. 
As manual annotation is expensive, we create synthetic native training data based on heuristic rules. 
We first extract 100M clean sentences from the WuDaoCorpora \cite{DBLP:journals/aiopen/YuanZDDLCZYT21}, which is mainly composed of articles crawled from native websites. 
Then, we inject errors into clean sentences by randomly replacing, inserting, deleting and swapping tokens. 
To better generate spelling errors, we also utilize confusion sets. 
The proportion of each error is set empirically. 
More details can be found in Appendix \ref{sec:pseudo}. 





\subsection{Experimental Results}
Table \ref{tab:data:exp} shows all experimental results. We evaluate models on the whole data of each domain.

In the \textsc{Media} and \textsc{Thesis} domains, the pseudo native training data significantly outperforms the real learner data, although the former is automatically crafted. This shows the text domain of training data can greatly influence model performance.

In the \textsc{Exam} domain, the real learner training data instead outperforms the pseudo native data substantially. 
We speculate the reason is that
most errors in the \textsc{Exam} domain are carefully designed to be difficult, which can hardly be simulated by simple rules but may occur in learner essays.

We also combine both data to make full use of them. We train our model on one kind of data until it converges, then continue to train it on another. As shown in the last two rows of Table \ref{tab:data:exp}, the data combinations lead to minor performance improvements in two domains, i.e., \textsc{Thesis} and  \textsc{Exam}.

Finally, the best F${_{0.5}}$ scores are 45.79, 31.87, and 20.02 for the \textsc{Media}, \textsc{Thesis}, and \textsc{Exam} domains, respectively, achieved by 3 different models.
It is worth noting that, although all models only have slight differences regarding overall average performance (the largest gap is just 1.72 F$_{0.5}$), they exhibit quite divergent behaviors in different domains (up to 13.72 F$_{0.5}$ gap).
This clearly demonstrates the value of NaSGEC as a multi-domain dataset to support a more comprehensive model evaluation.

\begin{table}[t]
\centering
\scalebox{0.9}{
\begin{small}
\begin{tabular}{@{~~}l@{~~}l@{~~}r@{~~}r@{~~}r@{~~}}
    \toprule
     & & \textbf{\textsc{Media}} & \textbf{\textsc{Thesis}} & \textbf{\textsc{Exam}} \\
    \midrule
    \parbox[t]{3mm}{\multirow{3}{*}{\rotatebox[origin=c]{90}{\textbf{Train}}}} & \textbf{\#Sent.} & 2,000 & 800 & 4,000  \\
     & \textbf{\#Err. Sent.} & 1,235 & 757 & 3,716 \\
     & \textbf{\#Ref.}  & 2,568 & 1,083 & 5,818 \\
    \midrule
    \parbox[t]{3mm}{\multirow{3}{*}{\rotatebox[origin=c]{90}{\textbf{Dev}}}} & \textbf{\#Sent.} & 500 & 200 & 1,000  \\
     & \textbf{\#Err. Sent.} & 312 & 141 & 723\\
     & \textbf{\#Ref.}  & 895 & 269 & 1,464 \\
    \midrule
    \parbox[t]{3mm}{\multirow{3}{*}{\rotatebox[origin=c]{90}{\textbf{Test}}}} & \textbf{\#Sent.} & 1,500 & 500 & 2,000  \\
     & \textbf{\#Err. Sent.} & 912 & 313 & 1,402   \\
     & \textbf{\#Ref.}  & 1,926 & 694 & 2,900  \\
    \bottomrule
\end{tabular}
\end{small}
}
\caption{Data split statistics of NaSGEC. 
}
\label{tab:split}
\end{table}

\section{Domain Analysis Within NaSGEC}

In this section, we conduct domain transfer experiments on NaSGEC by splitting data and performing fine-tuning.
We devise indicators of GEC domain shifts to gain more insights into the connections and differences between our domains.
To further improve model performance in specific domains, we also propose a simple domain-aware data augmentation method.

\subsection{Domain Transfer Experiments}
\label{sec:ft}

We perform domain transfer experiments by fine-tuning the baseline on training data from different domains. 
To facilitate fine-tuning, we split data into training/dev/test sets. The split statistics are listed in Table \ref{tab:split}.
For each domain, we select the best model
in it according to Table \ref{tab:data:exp}  as its baseline. After fine-tuning, we evaluate and compare all three fine-tuned models on this domain's test set.
All experimental results are presented in Table \ref{tab:transfer:1}.
We also perform error type analysis in Appendix \ref{sec:type:performance}.


\paragraph{In-domain results.} For in-domain results (fine-tune on one domain and evaluate on the same domain), we have the following observations.

First, the best performance in each domain is achieved by fine-tuning baselines on training data from the same domain, showing that in-domain data benefits more than out-of-domain data.
For example, although \textsc{Thesis}-train is much smaller than training sets in other domains, the \textsc{Thesis}-tuned model still performs best on \textsc{Thesis}-test. 

Second, fine-tuning models on little in-domain data can bring very significant performance improvements.
Specifically, in-domain fine-tuning leads to 10.89, 7.22, and 22.72 F$_{0.5}$ improvements in \textsc{Media}, \text{Thesis}, and \textsc{Exam}, respectively.

\paragraph{Out-of-domain results.} For out-of-domain results (fine-tune on one domain and evaluate on another), we have the following observations.

First, in the \textsc{Media} domain, fine-tuning the baseline with \textsc{Thesis}-train can lead to performance gain and vice versa, which indicates that the \textsc{Media} and \textsc{Thesis} domains are relatively similar.  

Second, in the \textsc{Exam} domain, fine-tuning with \textsc{Media}-train and \textsc{Thesis}-train both hurt the performance of the baseline. In turn, fine-tuning with \textsc{Exam}-train reduces the baseline performance in \textsc{Media} and  \textsc{Thesis}. This point to an obvious difference between \textsc{Exam} and the other 2 domains. 

\paragraph{Summary.}  Overall, fine-tuning models on training data from different domains leads to considerable performance changes, emphasizing the importance of \textit{domain} in GEC.
This also encourages us to study domain adaptation for GEC in the future.

\begin{table}[t]
\centering
\scalebox{0.65}{
\begin{tabular}{l|c|c|c}
\toprule
\textbf{Test $\rightarrow$}   & \textbf{\textsc{Media}} & \textbf{\textsc{Thesis}} & \textbf{\textsc{Exam}} \\
\textbf{Train $\downarrow$}  & \textbf{P/R/F$_{0.5}$} & \textbf{P/R/F$_{0.5}$} & \textbf{P/R/F$_{0.5}$} \\ \midrule

\textbf{\textit{Baseline}}  & 53.77/28.24/45.54      & 28.39/33.15/29.23       & 21.88/9.83/17.57   \\ \midrule
\textbf{\textsc{Media}}  & \textbf{61.35/42.72/56.43}     & 31.96/42.29/33.60       & 20.85/7.17/15.09      \\
\textbf{\textsc{Thesis}} & 52.65/33.40/47.21       & \textbf{34.96/43.96/36.45}     & 20.61/8.54/16.07  \\
\textbf{\textsc{Exam}}  & 49.16/24.74/41.06      & 27.93/31.58/28.59     & \textbf{48.29/24.23/40.29}   \\ \bottomrule
\end{tabular}
}
\caption{Results of transfer experiments on NaSGEC.}
\label{tab:transfer:1}
\end{table}



\begin{table*}[t]
\centering
\scalebox{0.8}{
\begin{tabular}{l|ccc|ccc|ccc}
\toprule
\textbf{Target $\rightarrow$}   & \multicolumn{3}{c|}{\textbf{\textsc{Media}-test}} & \multicolumn{3}{c|}{\textbf{\textsc{Thesis}-test}} & \multicolumn{3}{c}{\textbf{\textsc{Exam}-test}}  \\
 \textbf{Source $\downarrow$}  & \textbf{VO (\%) }     & \textbf{TDS}       & \textbf{EPO (\%) }   & \textbf{VO (\%) }     & \textbf{TDS}      & \textbf{EPO (\%) }   & \textbf{VO (\%) }      & \textbf{TDS}     & \textbf{EPO (\%) } \\
 \midrule
\textbf{\textsc{Media}-train}  & \textbf{65.03}     & \textbf{0.001}       & \textbf{25.84}   & 63.13    & 0.050    & 31.75            & 63.10     & 0.184      & 5.07    \\
\textbf{\textsc{Thesis}-train} & 56.47     & 0.025       & 22.77    & \textbf{75.73}       & \textbf{0.009}     & \textbf{33.05}    & 65.61   & 0.161      & 5.94     \\
\textbf{\textsc{Exam}-train}   & 62.97     &  0.210        & 6.94    & 66.33     &  0.139      & 10.29    & \textbf{68.30}     & \textbf{0.001}    & \textbf{14.89}   \\
\bottomrule
\end{tabular}
}
\caption{Vocabulary Overlap (VO), Type Distribution Similarity (TDS), and  Error Pattern Overlap (EPO) between training and test sets from different domains of NaSGEC. Specifically, VO and EPO are averaged over 3 calculations.}
\label{tab:domain:statistic}
\end{table*}

\subsection{Indicators of Domain Shifts}
\label{sec:indicator}

The domain transfer experiments reveal that there exist appreciable domain shifts in GEC. 
To better understand domain shifts in GEC, we further devise 3 indicators from a statistical perspective:

\begin{itemize}
    \item \textbf{Vocabulary Overlap (VO)} is defined as the ratio of the vocabulary of the target domain covered by the source domain. 
    Higher VO represents better vocabulary coverage.
    Since larger data usually covers vocabulary better, we sample 1,000 tokens from each domain when calculating VO to make it comparable.
    
    \item \textbf{Type Distribution Similarity (TDS)} is measured as the Kullback-Leibler (KL) divergence \cite{kullback1951information} between the error type distributions of two domains. The lower TDS indicates closer error type distributions. We extract and classify errors with the tool from MuCGEC \cite{DBLP:conf/naacl/0004LBLZLHZ22}.
    \item \textbf{Error Pattern Overlap (EPO)} is computed as the ratio of the error patterns in the target domain occurring in the source domain. We define an error pattern as a mapping from the erroneous span to the corresponding correct span. To eliminate the influence of data sizes, we randomly extract 300 edits from each domain to calculate EPO.
\end{itemize}


 We treat all 3 training sets as the source domains and all 3 test sets as the target domains. Then, we count the above indicators between them, as shown in Table \ref{tab:domain:statistic}. 
 With the help of these indicators, we revisit the results of domain transfer experiments and gain more insights, as shown below.
 
\paragraph{Explanation for in-domain results.}  In the previous section, we observe that using in-domain data for fine-tuning consistently outperforms out-of-domain data. Here, we find that the in-domain training sets best cover the vocabulary of the test sets, as reflected by VO. After looking at TDS and EPO, we also find that in-domain training sets have the  error distributions most similar to the test sets, in terms of both error types and patterns. These results show that different domains have their own characteristics in word selection and error distribution, which explains why using in-domain data contributes more than our-of-domain data.

\paragraph{Explanation for out-of-domain results.} Previously, we also observe that the \textsc{Media} and \textsc{Thesis} domains can benefit each other via fine-tuning, while the \textsc{Exam} domain is unable to help or get help from other domains.
From Table \ref{tab:domain:statistic}, we find that TDS/EPO is relatively low/high between \textsc{Media} and \textsc{Thesis}, exhibiting that these two domains have similar error distributions.
The reason can be that they are both built from realistic writing scenes, although \textsc{Media} is informal writing while \textsc{Thesis} is formal writing.

\begin{table}[t]
\centering
\scalebox{0.62}{
\begin{tabular}{l|ccc|ccc}
\toprule
& \multicolumn{3}{c|}{\textbf{\textsc{Media}}} & \multicolumn{3}{c}{\textbf{\textsc{Thesis}}} \\
  & \textbf{P} &  \textbf{R} & \textbf{F$_{0.5}$}        & \textbf{P} &  \textbf{R} & \textbf{F$_{0.5}$} \\ \midrule
\textbf{\textit{Pretrained Baseline}}  &  53.77 & 28.24 & 45.54 & 28.39 & 33.15 & 29.23      \\
\textbf{\hspace{0.3cm} + style adaptation}  &  54.31 & 29.79 & 46.63 & 29.09 & 34.91 & 30.09      \\
\textbf{\hspace{0.3cm} + error adaptation}  &  54.64 & 32.04 & 47.88 & 29.77 & 37.79 & 31.09      \\
\textbf{\hspace{0.3cm} + both}  & 57.29 & 32.41 & \textbf{49.66} & 31.17 & 43.17 & \textbf{33.00}  \\ \midrule
\textbf{\textit{Finetuned Baseline}}  &  61.35 & 42.72 & 56.43 & 34.96 & 43.96 & 36.45      \\
\textbf{\hspace{0.3cm} + style adaptation}  &  61.49 & 43.08 & 56.65 & 35.27 & 44.71 & 36.83      \\
\textbf{\hspace{0.3cm} + error adaptation}  &  61.72 & 43.65 & 57.00 & 35.12 & 45.30 & 36.77      \\
\textbf{\hspace{0.3cm} + both}  &  62.02 & 43.92 & \textbf{57.30} & 36.01 & 46.24 & \textbf{37.68}  \\   
\bottomrule
\end{tabular}
}
\caption{Results of domain-aware data augmentation.}
\label{tab:da}
\end{table}
As indicated by high TDS and low EPO compared to other domains, \textsc{Exam} has the most distinct error distribution. 
The possible reason is that  errors in \textsc{Exam} are deliberately designed to challenge native students and seldom occur in natives' daily writing. 
Such differences in error distribution can be strong evidence to explain the out-of-domain transfer phenomena.

\begin{table*}[t]
\centering
\scalebox{0.8}{
\begin{tabular}{l|ccc|ccc|ccc}
\toprule
\textbf{Target $\rightarrow$}   & \multicolumn{3}{c|}{\textbf{MuCGEC}}  & \multicolumn{3}{c|}{\textbf{CCTC}} & \multicolumn{3}{c}{\textbf{FCGEC}}  \\
 \textbf{Source $\downarrow$}   & \textbf{VO (\%) }      & \textbf{TDS}     & \textbf{EPO (\%) } & \textbf{VO (\%) }     & \textbf{TDS}       & \textbf{EPO (\%) }   & \textbf{VO (\%) }     & \textbf{TDS}      & \textbf{EPO (\%) }   \\
 \midrule
\textbf{\textsc{Media}}      & \textbf{72.50}   & \textbf{0.031}    & 5.79     & \textbf{64.43}     & \textbf{0.065}       & \textbf{42.26}       & 64.93     & 0.276      & 3.98   \\
\textbf{\textsc{Thesis}}  & 70.20     & 0.045       & 6.43  & 54.67      & 0.129    &  40.07    & 60.43   & 0.229      & 5.99       \\
\textbf{\textsc{Exam}}   & 70.03     &  0.078        & \textbf{7.31}    & 57.83     &  0.427      & 8.47    & \textbf{68.47}     & \textbf{0.010}    & \textbf{13.26}   \\
\bottomrule
\end{tabular}
}
\caption{Vocabulary Overlap (VO), Type Distribution Similarity (TDS), and  Error Pattern Overlap (EPO) from domains of NaSGEC to existing CGEC datasets. Specifically, VO and EPO are averaged over 3 calculations.}
\label{tab:domain:statistic:other}
\end{table*}

\begin{table}[t]
\centering
\scalebox{0.65}{
\begin{tabular}{l|c|c|c}
\toprule
\textbf{Test $\rightarrow$}  & \textbf{MuCGEC}  & \textbf{CCTC} & \textbf{FCGEC} \\
\textbf{Train $\downarrow$}  & \textbf{P/R/F$_{0.5}$} & \textbf{P/R/F$_{0.5}$} & \textbf{P/R/F$_{0.5}$} \\ \midrule

\textbf{\textit{Baseline}}     & 53.84/29.77/\textbf{46.34}  & 19.41/45.99/21.94     & 33.50/10.93/23.71      \\ \midrule
\textbf{\textsc{Media}}     & 52.67/21.88/41.10    & 20.88/55.40/\textbf{23.85}     & 32.07/5.12/15.62     \\
\textbf{\textsc{Thesis}}   &  60.61/21.09/44.09  & 17.98/55.73/20.80      & 34.10/8.15/20.83   \\
\textbf{\textsc{Exam}}     & 57.06/25.41/45.68  & 16.73/45.34/19.15      & 50.00/32.32/\textbf{45.07}    \\ \bottomrule
\end{tabular}
}
\caption{Results of transfer experiments from domains of NaSGEC to existing CGEC datasets.}
\label{tab:transfer}
\end{table}




\subsection{Domain-aware Data Augmentation}
As previously mentioned, the writing style and error distribution of the training data have a significant impact on the model's performance in a specific domain.
Hence, we propose a simple domain-aware data augmentation method by adapting the two aspects of pseudo data to the target domain.

We first perform the \textit{style adaptation}, which means using the raw data with a writing style similar to the target domain for augmentation.
For the \textsc{Media} domain, we collect 100k raw sentences from the \textit{Wechat public account platform}. For the \textsc{Thesis} domain, we collect 100k raw sentences from academic papers in the Chinese Scientific Literature (CSL) dataset \cite{li-etal-2021-csl}. We exclude \textsc{Exam} since it is difficult to gather sufficient raw data that comes from the same source.

We then conduct the \textit{error adaptation}. We inject 4 kinds of errors (missing, substituted, redundant, and word-order errors) to the raw sentence by rules and carefully control the error type distribution to simulate the target domain.

The experimental results are shown in Table \ref{tab:da}. The domain-aware data augmentation (+ both) leads to significant performance gains, even with the in-domain real training data (\textit{Finetuned Baseline}).
Only using either \textit{style adaptation} (+ style adaptation, without adjusting error type distribution) or \textit{error adaptation} (+ error adaptation, using 100k data from a general domain, i.e., WuDaoCorpora \cite{DBLP:journals/aiopen/YuanZDDLCZYT21}) still improves performance compared to the baseline, while the improvement is more marginal than simultaneously using both of them. Overall, this is a straightforward attempt, and we hope future work could study more methods for GEC domain adaptation based on NaSGEC.

\section{Comparison with Existing Datasets} 

In this section, we compare NaSGEC with existing CGEC datasets, including both native and learner datasets, 
by analysis of domain shift indicators (Table \ref{tab:domain:statistic:other}) and domain transfer experiments (Table \ref{tab:transfer}). Specifically, the baseline in Table \ref{tab:transfer} is trained with real learner data for MuCGEC and FCGEC, and pseudo native data for CCTC.


\paragraph{NaSGEC vs. Existing learner datasets.}
Most existing CGEC datasets are for learners. We select MuCGEC \cite{DBLP:conf/naacl/0004LBLZLHZ22} from them for comparison, because it actually covers several previous learner datasets, e.g., NLPCC18 \cite{zhao2018overview} and  CGED \cite{rao2018overview, rao2020overview}. 

From domain shift indicators in Table \ref{tab:domain:statistic:other}, we have two observations. 
First, VO is always high from our domains to MuCGEC, implying our data cover the vocabulary of MuCGEC well.
This may be because learners tend to use more common words. 
Second, all our domains get a mediocre level of TDS and EPO, revealing that errors made by native speakers differ from those made by learners. This illustrates why directly fine-tuning models on native data can not further boost performance on learner data.

From domain transfer experiments in Table \ref{tab:transfer}, we can see fine-tuning on domains of NaSGEC always results in performance degradation on MuCGEC, among them \textsc{Exam} brings the least decline.

We encourage future work to explore better ways to transfer between native and learner domains, which will allow us to apply the rich experience of learner GEC to under-explored native GEC.

\paragraph{NaSGEC vs. Existing native datasets.}
There are two existing native CGEC datasets, i.e., CCTC \cite{wang2022cctc} and FCGEC \cite{xu2022FCGEC}.

As shown in Table \ref{tab:domain:statistic:other}, CCTC is most like the \textsc{Media} domain of NaSGEC, possibly because they are both collected from natives' informal writing. EPO from \textsc{Media} and \textsc{Thesis} to CCTC is higher than 40\%, even exceeding their in-domain overlap ratios. As mentioned in Section \ref{sec:ana}, CCTC has a very high proportion of spelling errors. Spelling errors in Chinese, such as misusing ``的/地/得'', have fixed patterns and thus can be easily covered. In contrast, our data contains more long-tail and challenging grammatical errors.

Looking at transfer experiments, 
the recall of the baseline in CCTC greatly increased when fine-tuned on \textsc{Media} and \textsc{Thesis}, but the precision keeps low.
After carefully examining, we think this is due to the difference in error density.
As shown in Table \ref{tab:overall:statistic}, about 65.2\% and 70.0\% of sentences in \textsc{Media} and \textsc{Thesis} have errors, while the number in CCTC is just 9.3\%. Therefore, fine-tuning the baseline on our data will make it correct errors more aggressively, which causes poor precision in low error-density domains. In view of this, we hope future work can study how to transfer GEC models across domains with different error densities.

For FCGEC, fine-tuning the model on the \textsc{Exam} domain of NaSGEC leads to a huge improvement of over 22 F$_{0.5}$ scores, indicating they are highly compatible.
The indicator results also confirm this point.
We hope they can be two complementary resources to facilitate CGEC for Chinese teaching.

\section{Related Work}
\paragraph{Dataset.} Most GEC datasets are built for English. Early English GEC datasets, such as FCE \cite{yannakoudakis2011new}, NUCLE \cite{dahlmeier2013building}, and JFLEG \cite{napoles2017jfleg}, are built from student essays written by non-native English learners. After realizing the flaw of the limited text domain, researchers propose GMEG \cite{napoles2019enabling} and CWEB \cite{flachs2020grammatical}, 
two new datasets that broaden the target domain of English GEC to native speakers' daily writing.

Early CGEC work also primarily constructs datasets from learner essays, including NLPCC18 \cite{zhao2018overview}, CGED \cite{rao2018overview, rao2020overview}, YACLC \cite{wang2021yaclc}, and MuCGEC \cite{DBLP:conf/naacl/0004LBLZLHZ22}.
Concurrently with our work, some newly released CGEC datasets take native writing domains into account. CCTC \cite{wang2022cctc} annotates 1,500 web documents written by native speakers from the WuDaoCorpora \cite{DBLP:journals/aiopen/YuanZDDLCZYT21}. FCGEC \cite{xu2022FCGEC} mainly consists of sentences from multi-choice questions in Chinese examinations. Another work, NaCGEC \cite{Ma2022LinguisticRC}, collects data from Chinese examinations and news sites.


To the best of our knowledge, NaSGEC is the first CGEC dataset that annotates texts from multiple native domains under a unified scheme, which enables us to perform domain-wise experiments and analysis in CGEC for the first time.

\paragraph{Domain Adaptation.} Domain adaptation has been extensively studied in various NLP tasks \cite{DBLP:conf/coling/RamponiP20}, such as machine translation \cite{DBLP:conf/coling/ChuW18,DBLP:conf/acl/JiangLWZ20,DBLP:journals/tacl/PhamCY21}, syntax parsing \cite{li-etal-2019-semi-supervised-domain, yang-etal-2022-challenges}, and information extraction \cite{chen-qian-2021-bridge, lekhtman-etal-2021-dilbert}.


Compared with other fields, research on domain adaptation for GEC is under-explored.
Existing studies lie in adapting GEC models to a specific first language or proficiency level of the second language learners \cite{chollampatt-etal-2016-adapting, DBLP:conf/aclnut/NadejdeT19}. 
In this work, we build a multi-domain CGEC dataset from different writing scenarios and conduct basic cross-domain experiments, which can promote related research. We believe this is a valuable research direction for GEC even in the Large Language Model era \cite{fang2023chatgpt, coyne2023analysis, wu2023chatgpt, zhang2023multi}.



 
\section{Conclusion}
This paper presents NaSGEC, a new multi-domain native CGEC dataset, which consists of 12,500 sentences from three representative native domains. We clearly describe the construction process and perform detailed data analysis. We conduct benchmark experiments with the SOTA BART-based CGEC model and two kinds of training data. We also launch domain transfer experiments and devise domain shift indicators, in order to have a clearer understanding of our domains. 
We hope NaSGEC can spur future work on cross-domain GEC evaluation, domain adaptation for GEC, and more.
\section*{Limitations}

We think the limitations of our work are three-fold.
\begin{itemize}
    \item[(1)] As discussed in Section \ref{sec:dc}, we employ existing CGEC models to select sentences for annotation when building the \textsc{Media} and \textsc{Thesis} domains of NaSGEC. Although this reduces annotation costs, it inevitably introduces biases into our dataset. For instance, the proportion of complex syntax- or semantic-related errors may be lower than that in reality, since existing CGEC models fail to identify them. Note that although we manage to mitigate such biases by voting with multiple models, this issue still exists. Future work should explore how to automatically mine erroneous sentences from a low error-density domain with minimal biases. 
    \item[(2)] The current size of our dataset is relatively small. We will continuously collect more data from more diverse domains. Compared with other domains, \textsc{Thesis} has a much smaller data size (1.5k), as authorized papers are hard to obtain. In the future, we plan to cooperate with universities and thus accumulate more authorized data to enrich this domain. 
    \item[(3)] Based on our multi-domain NaSGEC, we have reported and analyzed cross-domain performance preliminarily. However, besides fine-tuning with small-scale data in the target domain, many other potentially helpful domain adaptation techniques can be tried.
    We believe cross-domain GEC is a valuable research topic and encourage future work to study it with NaSGEC.
\end{itemize}

\section*{Ethics Statement}

\paragraph{Data license.} For the \textsc{Exam} and \textsc{Media} domains of NaSGEC, we only collect sentences from public corpora or websites. For the \textsc{Thesis} domain, we have obtained permission from the authors of dissertations.  

\paragraph{Annotation payment.} During annotation, all annotators/reviewers were paid according to their finished task numbers and quality. The average salaries for annotators and reviewers are about 25 and 34 RMB per hour, respectively. 


\section*{Acknowledgements}
We thank all anonymous reviewers and the meta reviewer for their insightful comments, which will definitely help us improve this work in the future. This work was supported by the National Natural Science Foundation of China (Grant No. 62176173) and Alibaba Group through Alibaba Innovative  Research Program, and also supported by Project Funded by the Priority Academic Program Development of Jiangsu Higher Education Institutions.


\bibliography{anthology,custom}

\begin{thebibliography}{49}
\expandafter\ifx\csname natexlab\endcsname\relax\def\natexlab#1{#1}\fi

\bibitem[{Awasthi et~al.(2019)Awasthi, Sarawagi, Goyal, Ghosh, and
  Piratla}]{awasthi2019parallel}
Abhijeet Awasthi, Sunita Sarawagi, Rasna Goyal, Sabyasachi Ghosh, and Vihari
  Piratla. 2019.
\newblock \href {https://aclanthology.org/D19-1435/} {Parallel iterative edit
  models for local sequence transduction}.
\newblock In \emph{Proceedings of EMNLP-IJCNLP}, pages 4260--4270.

\bibitem[{Bryant et~al.(2019)Bryant, Felice, Andersen, and
  Briscoe}]{bryant2019bea}
Christopher Bryant, Mariano Felice, {\O}istein~E Andersen, and Ted Briscoe.
  2019.
\newblock \href {https://aclanthology.org/W19-4406/} {The {BEA}-2019 shared
  task on grammatical error correction}.
\newblock In \emph{Proceedings of BEA@ACL}, pages 52--75.

\bibitem[{Bryant et~al.(2022)Bryant, Yuan, Qorib, Cao, Ng, and
  Briscoe}]{bryant2022survey}
Christopher Bryant, Zheng Yuan, Muhammad~Reza Qorib, Hannan Cao, Hwee~Tou Ng,
  and Ted Briscoe. 2022.
\newblock \href {https://arxiv.org/abs/2211.05166} {Grammatical error
  correction: A survey of the state of the art}.
\newblock \emph{arXiv preprint arXiv:2211.05166}.

\bibitem[{Che et~al.(2010)Che, Li, and Liu}]{DBLP:conf/coling/CheLL10}
Wanxiang Che, Zhenghua Li, and Ting Liu. 2010.
\newblock \href {https://aclanthology.org/C10-3004/} {{LTP:} {A} {Chinese}
  language technology platform}.
\newblock In \emph{Proceedings of COLING}, pages 13--16.

\bibitem[{Chen and Qian(2021)}]{chen-qian-2021-bridge}
Zhuang Chen and Tieyun Qian. 2021.
\newblock \href {https://aclanthology.org/2021.acl-long.27/} {Bridge-based
  active domain adaptation for aspect term extraction}.
\newblock In \emph{Proceedings of ACL}, pages 317--327.

\bibitem[{Chollampatt et~al.(2016)Chollampatt, Hoang, and
  Ng}]{chollampatt-etal-2016-adapting}
Shamil Chollampatt, Duc~Tam Hoang, and Hwee~Tou Ng. 2016.
\newblock \href {https://aclanthology.org/D16-1195/} {Adapting grammatical
  error correction based on the native language of writers with neural network
  joint models}.
\newblock In \emph{Proceedings of EMNLP}, pages 1901--1911.

\bibitem[{Chu and Wang(2018)}]{DBLP:conf/coling/ChuW18}
Chenhui Chu and Rui Wang. 2018.
\newblock \href {https://aclanthology.org/C18-1111/} {A survey of domain
  adaptation for neural machine translation}.
\newblock In \emph{Proceedings of COLING}, pages 1304--1319.

\bibitem[{Coyne and Sakaguchi(2023)}]{coyne2023analysis}
Steven Coyne and Keisuke Sakaguchi. 2023.
\newblock \href {https://arxiv.org/abs/2303.14342} {An analysis of gpt-3's
  performance in grammatical error correction}.
\newblock \emph{arXiv preprint arXiv:2303.14342}.

\bibitem[{Dahlmeier et~al.(2013)Dahlmeier, Ng, and Wu}]{dahlmeier2013building}
Daniel Dahlmeier, Hwee~Tou Ng, and Siew~Mei Wu. 2013.
\newblock \href {https://aclanthology.org/W13-1703/} {Building a large
  annotated corpus of learner {English}: The nus corpus of learner {English}}.
\newblock In \emph{Proceedings of BEA@NAACL-HLT}, pages 22--31.

\bibitem[{Dai et~al.(2022)Dai, Li, Zhou, Feng, Zhao, Qiu, Li, and
  Tang}]{dai-etal-2022-whole}
Yong Dai, Linyang Li, Cong Zhou, Zhangyin Feng, Enbo Zhao, Xipeng Qiu, Piji Li,
  and Duyu Tang. 2022.
\newblock \href {https://aclanthology.org/2022.findings-acl.1/} {{``}{I}s whole
  word masking always better for {C}hinese {BERT}?{''}: Probing on {C}hinese
  grammatical error correction}.
\newblock In \emph{Proceedings of ACL (Short, Findings)}, pages 1--8.

\bibitem[{Fang et~al.(2023)Fang, Yang, Lan, Wong, Hu, Chao, and
  Zhang}]{fang2023chatgpt}
Tao Fang, Shu Yang, Kaixin Lan, Derek~F Wong, Jinpeng Hu, Lidia~S Chao, and Yue
  Zhang. 2023.
\newblock \href {https://arxiv.org/abs/2304.01746} {Is chatgpt a highly fluent
  grammatical error correction system? a comprehensive evaluation}.
\newblock \emph{arXiv preprint arXiv:2304.01746}.

\bibitem[{Flachs et~al.(2020)Flachs, Lacroix, Yannakoudakis, Rei, and
  S{\o}gaard}]{flachs2020grammatical}
Simon Flachs, Oph{\'e}lie Lacroix, Helen Yannakoudakis, Marek Rei, and Anders
  S{\o}gaard. 2020.
\newblock \href {https://aclanthology.org/2020.emnlp-main.680/} {Grammatical
  error correction in low error density domains: a new benchmark and analyses}.
\newblock In \emph{Proceedings of EMNLP}, pages 8467--8478.

\bibitem[{Jiang et~al.(2020)Jiang, Liang, Wang, and
  Zhao}]{DBLP:conf/acl/JiangLWZ20}
Haoming Jiang, Chen Liang, Chong Wang, and Tuo Zhao. 2020.
\newblock \href {https://aclanthology.org/2020.acl-main.165/} {Multi-domain
  neural machine translation with word-level adaptive layer-wise domain
  mixing}.
\newblock In \emph{Proceedings of ACL}, pages 1823--1834.

\bibitem[{Kingma and Ba(2014)}]{kingma2014adam}
Diederik~P Kingma and Jimmy Ba. 2014.
\newblock \href {https://arxiv.org/abs/1412.6980} {Adam: a method for
  stochastic optimization}.
\newblock \emph{arXiv preprint arXiv:1412.6980}.

\bibitem[{Kullback and Leibler(1951)}]{kullback1951information}
Solomon Kullback and Richard~A Leibler. 1951.
\newblock \href {https://www.jstor.org/stable/2236703} {On information and
  sufficiency}.
\newblock \emph{The annals of mathematical statistics}, 22(1):79--86.

\bibitem[{Lekhtman et~al.(2021)Lekhtman, Ziser, and
  Reichart}]{lekhtman-etal-2021-dilbert}
Entony Lekhtman, Yftah Ziser, and Roi Reichart. 2021.
\newblock \href {https://aclanthology.org/2021.emnlp-main.20/} {{DILBERT}:
  Customized pre-training for domain adaptation with category shift, with an
  application to aspect extraction}.
\newblock In \emph{Proceedings of EMNLP}, pages 219--230.

\bibitem[{Lewis et~al.(2020)Lewis, Liu, Goyal, Ghazvininejad, Mohamed, Levy,
  Stoyanov, and Zettlemoyer}]{lewis2020bart}
Mike Lewis, Yinhan Liu, Naman Goyal, Marjan Ghazvininejad, Abdelrahman Mohamed,
  Omer Levy, Veselin Stoyanov, and Luke Zettlemoyer. 2020.
\newblock \href {https://aclanthology.org/2020.acl-main.703.pdf} {{BART}:
  denoising sequence-to-sequence pre-training for natural language generation,
  translation, and comprehension}.
\newblock In \emph{Proceedings of ACL}, pages 7871--7880.

\bibitem[{Li et~al.(2022{\natexlab{a}})Li, Guo, Zhu, Sheng, Jiang, Ren, and
  Xu}]{DBLP:conf/aaai/LiGZSJRX22}
Jiquan Li, Junliang Guo, Yongxin Zhu, Xin Sheng, Deqiang Jiang, Bo~Ren, and
  Linli Xu. 2022{\natexlab{a}}.
\newblock \href {https://arxiv.org/abs/2205.10884} {Sequence-to-action:
  Grammatical error correction with action guided sequence generation}.
\newblock In \emph{Proceedings of AAAI}, pages 10974--10982.

\bibitem[{Li et~al.(2022{\natexlab{b}})Li, Zhang, Zhao, Shen, Weijie, Weiquan,
  and Hui}]{li-etal-2021-csl}
Yudong Li, Yuqing Zhang, Zhe Zhao, Linlin Shen, Liu Weijie, Mao Weiquan, and
  Zhang Hui. 2022{\natexlab{b}}.
\newblock \href {https://aclanthology.org/2022.coling-1.344/} {{CSL: A
  Large-scale {Chinese} Scientific Literature Dataset}}.
\newblock In \emph{Proceedings of COLING}, pages 3917--3923.

\bibitem[{Li et~al.(2019)Li, Peng, Zhang, Wang, and
  Si}]{li-etal-2019-semi-supervised-domain}
Zhenghua Li, Xue Peng, Min Zhang, Rui Wang, and Luo Si. 2019.
\newblock \href {https://aclanthology.org/P19-1229/} {Semi-supervised domain
  adaptation for dependency parsing}.
\newblock In \emph{Proceedings of ACL}, pages 2386--2395.

\bibitem[{Ma et~al.(2022)Ma, Li, Sun, Zhou, Huang, Zhang, Yangning, Liu, Li,
  Cao, Zheng, and Shen}]{Ma2022LinguisticRC}
Shirong Ma, Yinghui Li, Rongyi Sun, Qingyu Zhou, Shulin Huang, Dingchao Zhang,
  Li~Yangning, Ruiyang Liu, Zhongli Li, Yunbo Cao, Haitao Zheng, and Ying Shen.
  2022.
\newblock \href {https://aclanthology.org/2022.findings-emnlp.40} {Linguistic
  rules-based corpus generation for native {Chinese} grammatical error
  correction}.
\newblock In \emph{Proceedings of EMNLP (Findings)}, pages 576--589.

\bibitem[{Malmi et~al.(2019)Malmi, Krause, Rothe, Mirylenka, and
  Severyn}]{malmi2019encode}
Eric Malmi, Sebastian Krause, Sascha Rothe, Daniil Mirylenka, and Aliaksei
  Severyn. 2019.
\newblock \href {https://aclanthology.org/D19-1510/} {Encode, tag, realize:
  high-precision text editing}.
\newblock In \emph{Proceedings of EMNLP-IJCNLP}, pages 5054--5065.

\bibitem[{Nadejde and Tetreault(2019)}]{DBLP:conf/aclnut/NadejdeT19}
Maria Nadejde and Joel~R. Tetreault. 2019.
\newblock \href {https://aclanthology.org/D19-5504/} {Personalizing grammatical
  error correction: Adaptation to proficiency level and {L1}}.
\newblock In \emph{Proceedings of the 5th Workshop on Noisy User-generated
  Text, W-NUT@EMNLP}, pages 27--33.

\bibitem[{Napoles et~al.(2019)Napoles, N{\u{a}}dejde, and
  Tetreault}]{napoles2019enabling}
Courtney Napoles, Maria N{\u{a}}dejde, and Joel Tetreault. 2019.
\newblock \href {https://aclanthology.org/Q19-1032/} {Enabling robust
  grammatical error correction in new domains: data sets, metrics, and
  analyses}.
\newblock \emph{TACL}, 7:551--566.

\bibitem[{Napoles et~al.(2017)Napoles, Sakaguchi, and
  Tetreault}]{napoles2017jfleg}
Courtney Napoles, Keisuke Sakaguchi, and Joel Tetreault. 2017.
\newblock \href {https://aclanthology.org/E17-2037/} {{JFLEG}: a fluency corpus
  and benchmark for grammatical error correction}.
\newblock In \emph{Proceedings of EACL}, pages 229--234.

\bibitem[{Omelianchuk et~al.(2020)Omelianchuk, Atrasevych, Chernodub, and
  Skurzhanskyi}]{omelianchuk2020gector}
Kostiantyn Omelianchuk, Vitaliy Atrasevych, Artem Chernodub, and Oleksandr
  Skurzhanskyi. 2020.
\newblock \href {https://aclanthology.org/2020.bea-1.16/}
  {{GECToR}--grammatical error correction: tag, not rewrite}.
\newblock In \emph{Proceedings of BEA@ACL}, pages 163--170.

\bibitem[{Ott et~al.(2019)Ott, Edunov, Baevski, Fan, Gross, Ng, Grangier, and
  Auli}]{ott2019fairseq}
Myle Ott, Sergey Edunov, Alexei Baevski, Angela Fan, Sam Gross, Nathan Ng,
  David Grangier, and Michael Auli. 2019.
\newblock \href {https://aclanthology.org/N19-4009/} {fairseq: A fast,
  extensible toolkit for sequence modeling}.
\newblock In \emph{Proceedings of NAACL-HLT(Demo)}, pages 48--53.

\bibitem[{Pham et~al.(2021)Pham, Crego, and Yvon}]{DBLP:journals/tacl/PhamCY21}
Minh~Quang Pham, Josep~Maria Crego, and Fran{\c{c}}ois Yvon. 2021.
\newblock \href {https://aclanthology.org/2021.tacl-1.2/} {Revisiting
  multi-domain machine translation}.
\newblock \emph{TACL}, 9:17--35.

\bibitem[{Ramponi and Plank(2020)}]{DBLP:conf/coling/RamponiP20}
Alan Ramponi and Barbara Plank. 2020.
\newblock \href {https://aclanthology.org/2020.coling-main.603/} {Neural
  unsupervised domain adaptation in {NLP} - {A} survey}.
\newblock In \emph{Proceedings of COLING}, pages 6838--6855.

\bibitem[{Rao et~al.(2018)Rao, Gong, Zhang, and Xun}]{rao2018overview}
Gaoqi Rao, Qi~Gong, Baolin Zhang, and Endong Xun. 2018.
\newblock \href {https://aclanthology.org/W18-3706/} {Overview of {NLPTEA}-2018
  share task {{Chinese}} grammatical error diagnosis}.
\newblock In \emph{Proceedings of NLPTEA@ACL}, pages 42--51.

\bibitem[{Rao et~al.(2020)Rao, Yang, and Zhang}]{rao2020overview}
Gaoqi Rao, Erhong Yang, and Baolin Zhang. 2020.
\newblock \href {https://aclanthology.org/2020.nlptea-1.4/} {Overview of
  {NLPTEA}-2020 shared task for {{Chinese}} grammatical error diagnosis}.
\newblock In \emph{Proceedings of NLPTEA@AACL}, pages 25--35.

\bibitem[{Richards(1987)}]{richards1987type}
Brian Richards. 1987.
\newblock \href
  {https://www.cambridge.org/core/journals/journal-of-child-language/article/abs/typetoken-ratios-what-do-they-really-tell-us/B15717A4D91390ED7E2F2DA143BA1DDB}
  {Type/token ratios: What do they really tell us?}
\newblock \emph{Journal of Child Language}, 14(2):201--209.

\bibitem[{Sakaguchi et~al.(2016)Sakaguchi, Napoles, Post, and
  Tetreault}]{sakaguchi2016reassessing}
Keisuke Sakaguchi, Courtney Napoles, Matt Post, and Joel Tetreault. 2016.
\newblock \href {https://aclanthology.org/Q16-1013/} {Reassessing the goals of
  grammatical error correction: fluency instead of grammaticality}.
\newblock \emph{TACL}, 4:169--182.

\bibitem[{Shao et~al.(2021)Shao, Geng, Liu, Dai, Yang, Zhe, Bao, and
  Qiu}]{shao2021cpt}
Yunfan Shao, Zhichao Geng, Yitao Liu, Junqi Dai, Fei Yang, Li~Zhe, Hujun Bao,
  and Xipeng Qiu. 2021.
\newblock \href {https://arxiv.org/abs/2109.05729} {{CPT}: a pre-trained
  unbalanced transformer for both {{Chinese}} language understanding and
  generation}.
\newblock \emph{arXiv preprint arXiv:2109.05729}.

\bibitem[{Szegedy et~al.(2016)Szegedy, Vanhoucke, Ioffe, Shlens, and
  Wojna}]{szegedy2016rethinking}
Christian Szegedy, Vincent Vanhoucke, Sergey Ioffe, Jon Shlens, and Zbigniew
  Wojna. 2016.
\newblock \href {https://ieeexplore.ieee.org/document/7780677} {Rethinking the
  inception architecture for computer vision}.
\newblock In \emph{Proceedings of ICCV}, pages 2818--2826.

\bibitem[{Wang et~al.(2022)Wang, Duan, Wu, Che, Chen, and Hu}]{wang2022cctc}
Baoxin Wang, Xingyi Duan, Dayong Wu, Wanxiang Che, Zhigang Chen, and Guoping
  Hu. 2022.
\newblock \href {https://aclanthology.org/2022.coling-1.294/} {{CCTC}: A
  cross-sentence {Chinese} text correction dataset for native speakers}.
\newblock In \emph{Proceedings of COLING}, pages 3331--3341.

\bibitem[{Wang et~al.(2021)Wang, Kong, Yang, Wang, Lu, Hu, He, Liu, Chen, Yang,
  and Sun}]{wang2021yaclc}
Yingying Wang, Cunliang Kong, Liner Yang, Yijun Wang, Xiaorong Lu, Renfen Hu,
  Shan He, Zhenghao Liu, Yun Chen, Erhong Yang, and Maosong Sun. 2021.
\newblock \href {https://arxiv.org/abs/2112.15043} {{YACLC}: a {{Chinese}}
  learner corpus with multidimensional annotation}.
\newblock \emph{arXiv preprint arXiv:2112.15043}.

\bibitem[{Wu et~al.(2023)Wu, Wang, Wan, Jiao, and Lyu}]{wu2023chatgpt}
Haoran Wu, Wenxuan Wang, Yuxuan Wan, Wenxiang Jiao, and Michael Lyu. 2023.
\newblock \href {https://arxiv.org/abs/2303.13648} {Chatgpt or grammarly?
  evaluating chatgpt on grammatical error correction benchmark}.
\newblock \emph{arXiv preprint arXiv:2303.13648}.

\bibitem[{Wu and Wu(2022)}]{wu2022from}
Xiuyu Wu and Yunfang Wu. 2022.
\newblock \href {https://arxiv.org/abs/2211.01625} {From spelling to grammar:
  {A} new framework for {Chinese} grammatical error correction}.
\newblock \emph{ArXiv}, abs/2211.01625.

\bibitem[{Xu et~al.(2022)Xu, Wu, Peng, Fu, and Cai}]{xu2022FCGEC}
Lvxiaowei Xu, Jian-Cheng Wu, Jiawei Peng, Jiayu Fu, and Ming Cai. 2022.
\newblock \href {https://aclanthology.org/2022.findings-emnlp.137} {{FCGEC}:
  Fine-grained corpus for {Chinese} grammatical error correction}.
\newblock In \emph{Proceedings of EMNLP (Findings)}, pages 1900--1918.

\bibitem[{Yang et~al.(2022)Yang, Cui, Ning, Wu, and
  Zhang}]{yang-etal-2022-challenges}
Sen Yang, Leyang Cui, Ruoxi Ning, Di~Wu, and Yue Zhang. 2022.
\newblock \href {https://aclanthology.org/2022.findings-acl.11/} {Challenges to
  open-domain constituency parsing}.
\newblock In \emph{Proceedings of ACL (Findings)}, pages 112--127.

\bibitem[{Yannakoudakis et~al.(2011)Yannakoudakis, Briscoe, and
  Medlock}]{yannakoudakis2011new}
Helen Yannakoudakis, Ted Briscoe, and Ben Medlock. 2011.
\newblock \href {https://aclanthology.org/P11-1019/} {A new dataset and method
  for automatically grading {ESOL} texts}.
\newblock In \emph{Proceedings of ACL}, pages 180--189.

\bibitem[{Yuan et~al.(2021)Yuan, Zhao, Du, Ding, Liu, Cen, Zou, Yang, and
  Tang}]{DBLP:journals/aiopen/YuanZDDLCZYT21}
Sha Yuan, Hanyu Zhao, Zhengxiao Du, Ming Ding, Xiao Liu, Yukuo Cen, Xu~Zou,
  Zhilin Yang, and Jie Tang. 2021.
\newblock \href
  {https://www.sciencedirect.com/science/article/pii/S2666651021000152}
  {{WuDaoCorpora}: {A} super large-scale {Chinese} corpora for pre-training
  language models}.
\newblock \emph{{AI} Open}, 2:65--68.

\bibitem[{Zhang(2009)}]{zhang2009hsk}
Baolin Zhang. 2009.
\newblock \href
  {http://www.en.cnki.com.cn/Article_en/CJFDTotal-GHJY200904011.htm} {Features
  and functions of the {HSK} dynamic composition corpus (in {{Chinese}})}.
\newblock \emph{International {Chinese} Language Education}, 4:71--79.

\bibitem[{Zhang et~al.(2023)Zhang, Cui, Cai, Huang, Fang, and
  Bi}]{zhang2023multi}
Yue Zhang, Leyang Cui, Deng Cai, Xinting Huang, Tao Fang, and Wei Bi. 2023.
\newblock \href {https://arxiv.org/abs/2305.13225} {Multi-task instruction
  tuning of llama for specific scenarios: A preliminary study on writing
  assistance}.
\newblock \emph{arXiv preprint arXiv:2305.13225}.

\bibitem[{Zhang et~al.(2022{\natexlab{a}})Zhang, Li, Bao, Li, Zhang, Li, Huang,
  and Zhang}]{DBLP:conf/naacl/0004LBLZLHZ22}
Yue Zhang, Zhenghua Li, Zuyi Bao, Jiacheng Li, Bo~Zhang, Chen Li, Fei Huang,
  and Min Zhang. 2022{\natexlab{a}}.
\newblock \href {https://aclanthology.org/2022.naacl-main.227/} {{MuCGEC}: a
  multi-reference multi-source evaluation dataset for {Chinese} grammatical
  error correction}.
\newblock In \emph{Proceedings of NAACL-HLT}, pages 3118--3130.

\bibitem[{Zhang et~al.(2022{\natexlab{b}})Zhang, Zhang, Li, Bao, Li, and
  Zhang}]{zhang2022syngec}
Yue Zhang, Bo~Zhang, Zhenghua Li, Zuyi Bao, Chen Li, and Min Zhang.
  2022{\natexlab{b}}.
\newblock \href {https://aclanthology.org/2022.emnlp-main.162} {{SynGEC}:
  Syntax-enhanced grammatical error correction with a tailored gec-oriented
  parser}.
\newblock In \emph{Proceedings of EMNLP}, pages 2518--2531.

\bibitem[{Zhao et~al.(2019)Zhao, Wang, Shen, Jia, and
  Liu}]{DBLP:conf/naacl/ZhaoWSJL19}
Wei Zhao, Liang Wang, Kewei Shen, Ruoyu Jia, and Jingming Liu. 2019.
\newblock \href {https://aclanthology.org/N19-1014/} {Improving grammatical
  error correction via pre-training a copy-augmented architecture with
  unlabeled data}.
\newblock In \emph{Proceedings of NAACL-HLT}, pages 156--165.

\bibitem[{Zhao et~al.(2018)Zhao, Jiang, Sun, and Wan}]{zhao2018overview}
Yuanyuan Zhao, Nan Jiang, Weiwei Sun, and Xiaojun Wan. 2018.
\newblock \href
  {https://link.springer.com/chapter/10.1007/978-3-319-99501-4_41} {Overview of
  the {NLPCC} 2018 shared task: grammatical error correction}.
\newblock In \emph{CCF International Conference on Natural Language Processing
  and {Chinese} Computing (NLPCC)}, pages 439--445.

\end{thebibliography}
\bibliographystyle{acl_natbib}

\appendix


\section{Annotation Tool}
\label{sec:anno:tool}

\begin{figure}[h!]
\centering
\includegraphics[scale=0.3]{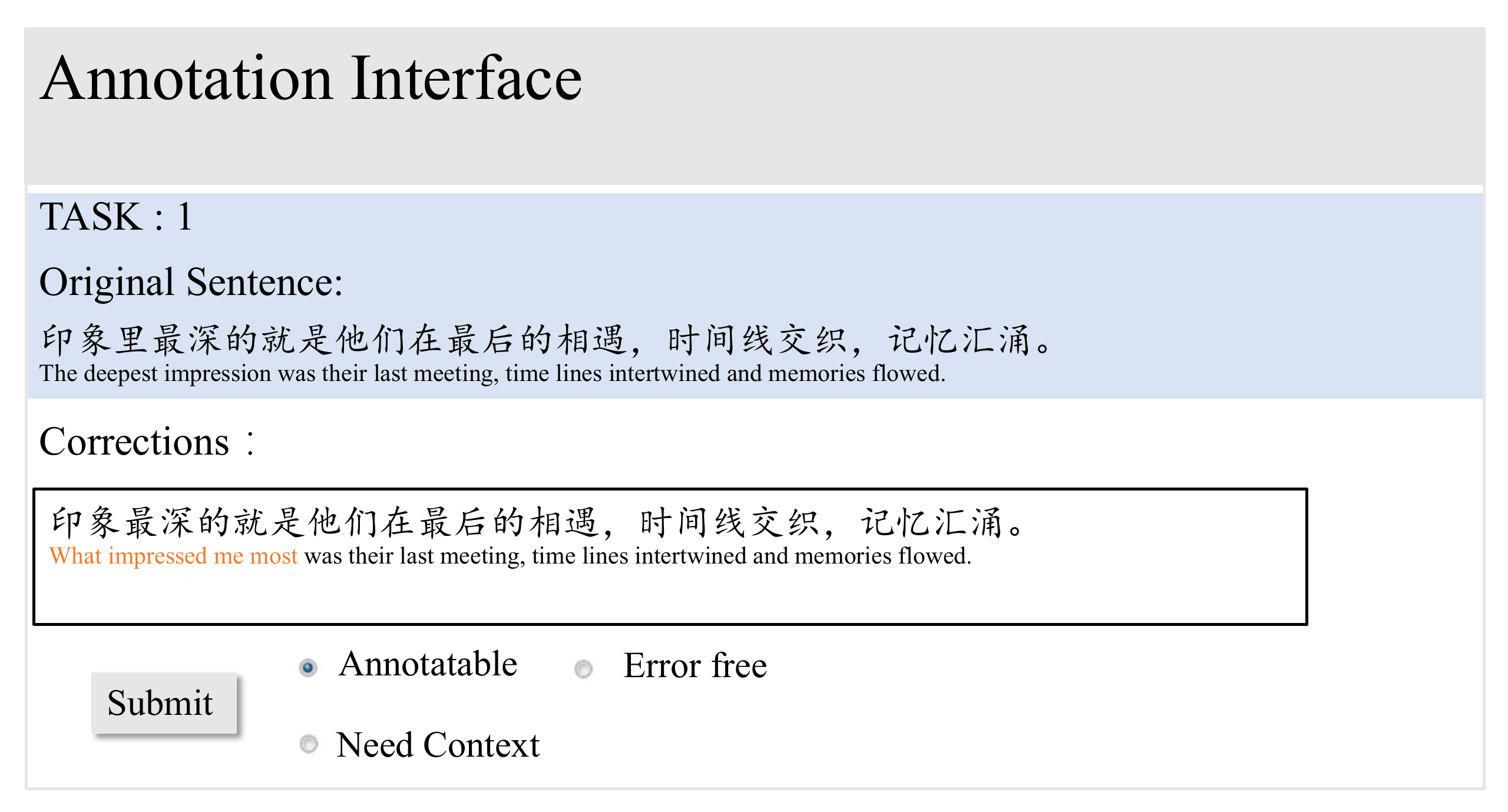}
\caption{Our annotation interface.}
\label{fig:anno:inter}
\end{figure}

We present the annotation interface of our annotation tool in Figure \ref{fig:anno:inter}. Given a potentially erroneous sentence, the annotator can rewrite it in a text box if he/she finds this sentence contains errors. If the sentence is correct, the annotator can directly click the \texttt{Error Free} button and submit. 

Specifically, when annotating the \textsc{Media} and \textsc{Thesis} domains, we provide annotators with the context of each sentence. Because sentences in these domains are extracted from complete essays or dissertations, they may need cross-sentence information to correct. 
We ask our annotators to mark such sentences with the \texttt{Need Context} button to facilitate future study in document-level CGEC.

\begin{figure}[h]
\centering
\includegraphics[scale=0.3]{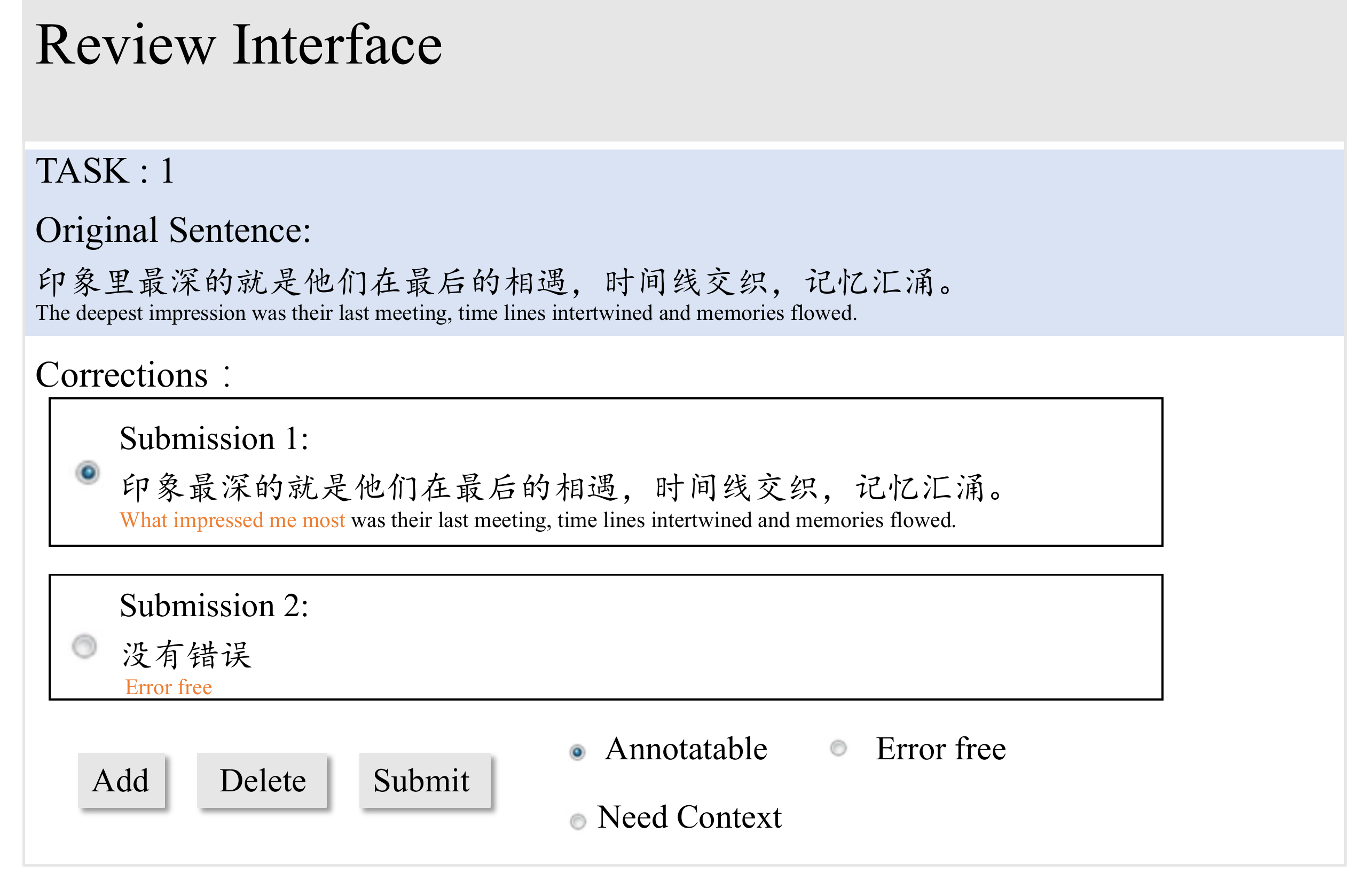}
\caption{Our review interface.}
\label{fig:review:inter}
\end{figure}

Figure \ref{fig:review:inter} shows our review interface. The reviewer can choose whether accept each submission by clicking the check box before it. Considering other valid answers may be missed by annotators, the reviewer can also click the \texttt{Add} button to input a new correction for supplementary.

\section{Experimental Details}
\label{sec:exp:detail}
\begin{table}[h!]
\centering
\scalebox{0.72}{
\begin{tabular}{lc}
\hline
\textbf{Hyper-parameter}      & \textbf{Value}                        \\ \hline
\multicolumn{2}{c}{\textbf{Training}}                  \\ \hline
Pretrained Language model   & \begin{tabular}[c]{@{}c@{}}Chinese-BART-large \\ \citep{shao2021cpt} \end{tabular}                        \\
Update steps   & 200,000                           \\
Devices            & 8 Tesla V100 GPU (32GB)      \\
Batch size per GPU & 8096 tokens                        \\
Optimizer          & \begin{tabular}[c]{@{}c@{}}Adam \citep{kingma2014adam}\\ ($\beta_1=0.9,\beta_2=0.98,\epsilon=1 \times 10^{-8}$) \end{tabular}                        \\
Learning rate      &  $3 \times 10^{-5}$                          \\
Warmup updates             & 4000                         \\
Max length  &   128                            \\
Loss function      & \begin{tabular}[c]{@{}c@{}}Label smoothed cross entropy \\ (label-smoothing=0.1)\\\citep{szegedy2016rethinking} \end{tabular}  \\
Dropout            &  0.3                        \\ 
Dropout-src            & 0.2                          \\ 
\hline
\multicolumn{2}{c}{\textbf{Fine-tuning}}                   \\ \hline
Devices            & 1 Tesla V100 GPU (32GB)      \\
Max epochs   & 100                           \\
Learning rate      & $1 \times 10^{-5}$                         \\
Batch size per GPU & 1024 tokens                        \\
\hline
\multicolumn{2}{c}{\textbf{Generation}}                    \\ \hline
Beam size          & 12                           \\
Max input length  &  128                           \\
\hline
\end{tabular}
}
\caption{Our hyper-parameter settings.}
\label{tab:hp}
\end{table}
We use the \texttt{fairseq} toolkit\footnote{\url{https://github.com/facebookresearch/fairseq}} \cite{ott2019fairseq} to build our benchmark models.
Our model is based on the large variant of the Chinese BART \cite{shao2021cpt}\footnote{\url{https://huggingface.co/fnlp/bart-large-chinese}}, which has about 400M parameters.
Following \citet{zhang2022syngec}, we extend the original vocabulary of the Chinese BART to cover some common but missed Chinese characters and punctuation, e.g., Chinese quotation marks, which they find can greatly improve model performance. 

We list detailed experimental hyper-parameter settings in Table \ref{tab:hp}. The total training time for using real learner data (about 1.35M sentence pairs) is about 10 hours. The total training time for using pseudo native data (about 100M sentence pairs) is about 7 days. Due to the limitation of time and computation resources, the benchmark results in Table \ref{tab:data:exp} are reported over a single run. The fine-tuning time is about 20 minutes. All fine-tuning results in Table \ref{tab:transfer:1} and Table \ref{tab:transfer} are averaged over 3 runs with distinct random seeds.


\section{Results of the Seq2Edit Model}

\label{sec:seq2edit}
Besides Seq2Seq-based models like BART \cite{lewis2020bart}, there is another competitive CGEC paradigm called Seq2Edit. The Seq2Edit-based models first predict a sequence of edits, and then apply them to the erroneous sentence to conduct corrections \cite{malmi2019encode, awasthi2019parallel}. Recently, \citet{DBLP:conf/naacl/0004LBLZLHZ22} adapt GECToR \cite{omelianchuk2020gector}, a widely-used Seq2Edit model in English, to Chinese and find it can achieve promising performance. Hence, we follow their efforts and test the ability of Chinese GECToR on NaSGEC, as shown in Table \ref{tab:seq2edit}. Both BART and GECToR are trained on real learner training data described in Section \ref{sec:train:data}.

We can see that, in \textsc{Media} and \textsc{Exam}, Seq2Seq outperforms Seq2Edit substantially. However, in \textsc{Thesis}, Seq2Edit performs significantly better. We attribute this to Seq2Edit's natural ability to copy. Seq2Edit can directly copy tokens from the source sentence by predicting the \texttt{Keep} tag. In \textsc{Thesis}, there are many English words and technical terms, which Seq2Seq tends to mis-correct while Seq2Edit keeps unchanged. So Seq2Edit achieves a much higher precision in this domain. In view of this, we plan to enhance our BART-based benchmark models with the copy mechanism \cite{DBLP:conf/naacl/ZhaoWSJL19} or other approaches in the future.

\textbf{
\begin{table}[t!]
\centering
\scalebox{1}{
\begin{tabular}{lccc}
\toprule
\textbf{\textsc{Media}} & \textbf{P} & \textbf{R} & \textbf{F$_{0.5}$} \\ \midrule
\textbf{BART}    & \textbf{35.96} & \textbf{29.15} & \textbf{34.35}  \\
\textbf{GECToR}     & 33.36 & 19.85 & 29.36 \\ \midrule
\textbf{\textsc{Thesis}} & \textbf{P} & \textbf{R} & \textbf{F$_{0.5}$} \\ \midrule
\textbf{BART} & 24.16 & \textbf{34.06} & 25.65 \\
\textbf{GECToR} &  \textbf{42.29} & 18.20 & \textbf{33.44} \\ \midrule
\textbf{\textsc{Exam}} & \textbf{P} & \textbf{R} & \textbf{F$_{0.5}$} \\ \midrule
\textbf{BART}      & \textbf{23.01} & \textbf{11.31} & \textbf{19.06}  \\
\textbf{GECToR}     & 20.93 & 8.80 & 16.41 \\ \bottomrule
\end{tabular}
}
\caption{Experimental results of the Seq2Edit-based model (GECToR) compared with the Seq2Seq-based model (BART) on NaSGEC.}
\label{tab:seq2edit}
\end{table}}

\begin{figure*}[h]
\centering
\includegraphics[scale=0.57]{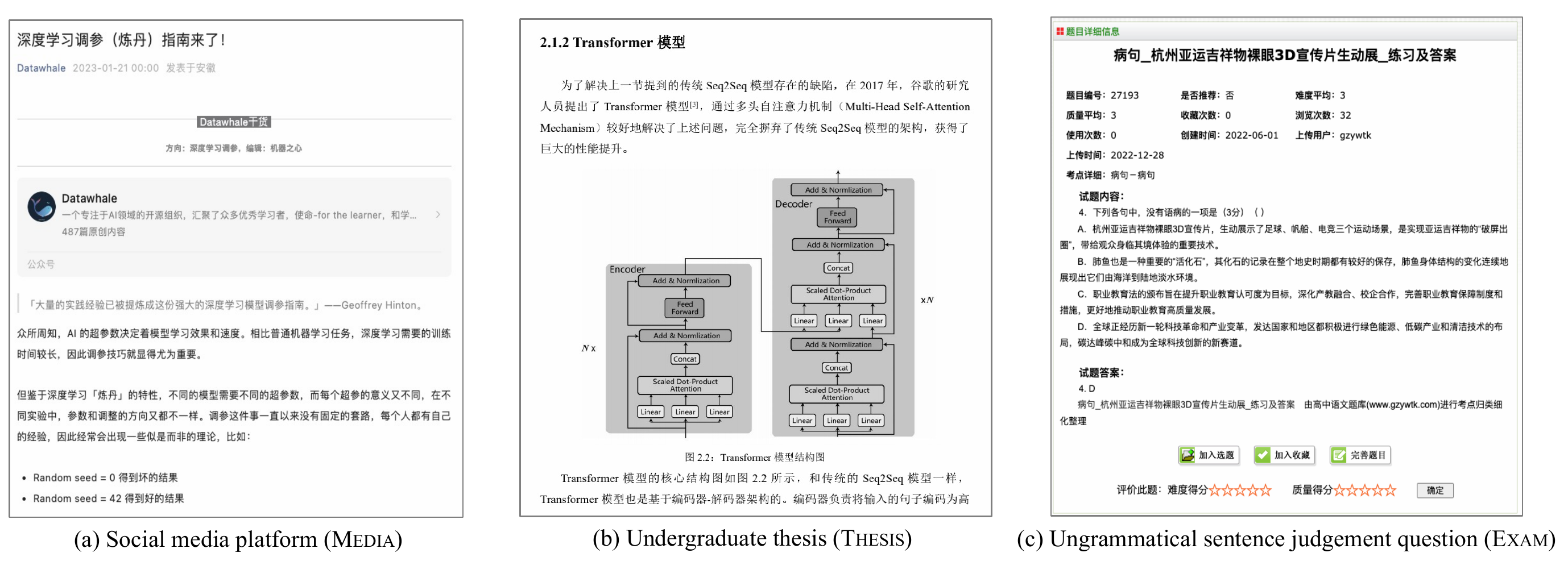}
\caption{The screenshots of data sources for our 3 domains.}
\label{fig:data:source}
\end{figure*}

\section{Pseudo Data Generation}
\label{sec:pseudo}
We use rule-based corruption to generate large-scale synthetic training data from clean native corpora. Specifically, we randomly select 100M sentences from the WuDao corpora \cite{DBLP:journals/aiopen/YuanZDDLCZYT21}\footnote{\url{https://data.wudaoai.cn/home}} as the seed corpus, which is mainly composed of website articles written by native speakers. We select tokens for corruption with a probability of 0.05 and perform the following operations with corresponding probabilities (in parentheses):
\begin{itemize}
    \item \textbf{Replacement} (0.55): We replace the current token with another token in its confusion set (0.5) or a random token from the whole vocabulary (0.5).
    \item \textbf{Insertion} (0.2): We insert the same token (0.5) or a random token from the whole vocabulary (0.5) before the current token 
    \item \textbf{Deletion} (0.2): We delete the current token.
    \item \textbf{Swap} (0.05): We swap the current token and the token after it.
\end{itemize}

Following \citet{dai-etal-2022-whole}, we inject noises from both character and word granularity to achieve better performance, which means each sentence is segmented into either the word (0.5) or character (0.5) sequence before corruption. The word-level and character-level confusion sets are built considering phonetics and glyphs.

\begin{figure}[tb!] 
\centering 
\scalebox{0.7}{
\begin{tikzpicture}
\begin{axis}[
    legend style={
                fill opacity=2,
                legend cell align={left},
                text=black,
                at={(0.68,0.50)},
                anchor=north west,
                font=\small,
              },
    xlabel={\textbf{Data size (M)}},
    ylabel={\textbf{F$_{0.5}$}},
    ylabel style = {yshift=-10pt},
    ymin=0, ymax=50,
    symbolic x coords={1, 10, 50, 100},
    xtick=data,
    ymajorgrids=true,
    xmajorgrids=true,
    grid style=dashed,
]

\addplot[
    color=brickred!80,thick,
    mark=square*,
    mark options={solid,mark size=2pt}
    ]
    coordinates {
    (1,35.12)(10,42.48)(50,44.44)(100,45.79)
    };
    \addlegendentry{\textsc{Media}}

\addplot[
    color=midnightblue!80, thick,
    mark=triangle*,
    mark options={solid,mark size=2pt}
    ]
    coordinates {
    (1,25.68)(10,29.22)(50,30.66)(100,31.15)
    };
    \addlegendentry{\textsc{Thesis}}

\addplot[
    color=burntorange!80,thick,
    mark=diamond*,
    mark options={solid,mark size=2pt}
    ]
    coordinates {
    (1,5.78)(10,5.96)(50,6.25)(100,6.30)
    };
    \addlegendentry{\textsc{Exam}}

\end{axis}
\end{tikzpicture}
}

\caption{
Impact of pseudo data size in different domains of NaSGEC.
}
\label{fig:data:num}
\end{figure}
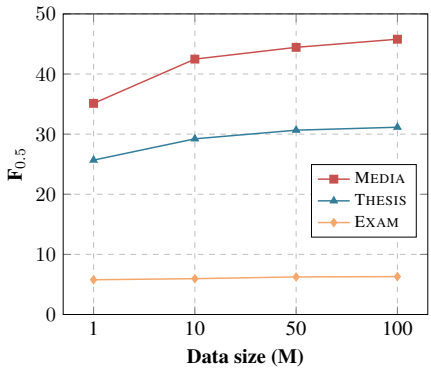

We also show the effect of the size of pseudo data in Figure \ref{fig:data:num}. When the data size increases, the model performance continuously improves in the \textsc{Media} and \textsc{Thesis} domains, whereas the model performance in the \textsc{Exam} domain keeps low.

\section{Error Type Performance}

\label{sec:type:performance}
In Table \ref{tab:type:performance}, we evaluate the error type performance of each domain's best model on NaSGEC. The best model denotes the fine-tuned model achieving the highest F$_{0.5}$ score  in Table \ref{tab:transfer:1}.

In all domains, models repair redundant errors consistently well, as their corrections do not need to generate new content and are the easiest and most deterministic. 
In contrast, models encounter difficulties in handling word-order errors universally since such errors require long-range structural knowledge to correct.

In terms of substituted and missing errors, models exhibit divergent behaviours. The performance on substituted errors in \textsc{Media} is very high, probably because they are often spelling and punctuation errors. However, as another realistic writing scene, \textsc{Thesis} has a much inferior performance on substituted errors due to the low correction precision. After studying cases, we find \textsc{Thesis} contains many English words (e.g., LSTM) and technical terms (e.g., 支持向量机, \textit{supporting vector machine}), which usually cause miscorrection. Besides, the performance on substituted errors in \textsc{Exam} is also quite low, owing to their complexity.

Considering missing errors, the model performs much better in \textsc{Media} than others. As discussed before, we observe that a large proportion of missing errors in \textsc{Media} is caused by missing punctuation, which well-trained models can easily handle.

\begin{table}[!t]
\centering
\scalebox{0.7}{
\begin{tabular}{lcccc}
\toprule
& \textbf{\textsc{Media}} & \textbf{\textsc{Thesis}} & \textbf{\textsc{Exam}} \\ 
& \textbf{P/R/F$_{0.5}$} & \textbf{P/R/F$_{0.5}$} & \textbf{P/R/F$_{0.5}$}  \\ 
 \hline
 \textbf{S} & 59.91/51.66/58.06 & 29.79/60.64/33.17 &  25.38/15.07/22.33  \\
\textbf{M} & 67.56/32.54/55.59 & 47.37/15.38/33.46 & 44.21/19.62/35.35 \\
\textbf{R} & 59.41/42.44/55.01 & 65.71/34.85/55.83 & 66.10/41.42/59.06 \\
\textbf{W} & 40.00/12.77/28.04 & 42.25/12.75/28.88 & 29.74/9.46/20.82  \\
\bottomrule
\end{tabular}
}
\caption{The fine-grained performance of each domain's best model regarding error types. S: Substituted errors, M: Missing errors, R: Redundant errors, W: Word-order errors.
} 
\label{tab:type:performance}
\end{table}
\begin{table*}[h!]
\centering
\scalebox{0.73}{
\begin{tabular}{ll}
\toprule
\multicolumn{2}{c}{\textbf{Domain: \textsc{Media}}} \\ \hline
\textbf{Source} & \begin{tabular}[c]{@{}l@{}}30日下午齐鲁晚报的一名读者报料称，南湖镇两个女孩泥水，正在医院抢救。\\ 
On the afternoon of the 30th a reader of the Qilu Evening News reported that two girls in Nanhu Town muddy water, \\ and were being rescued in the hospital. \end{tabular} \\ 
\textbf{Ref. 1} & \begin{tabular}[c]{@{}l@{}}30日下午\textcolor{red}{，}齐鲁晚报的一名读者报料称，南湖镇两个女孩\textcolor{red}{溺水}，正在医院抢救。\\ 
On the afternoon of the 30th\textcolor{red}{,} a reader of the Qilu Evening News reported that two girls in Nanhu Town \textcolor{red}{drowned}, \\ and were being rescued in the hospital. \end{tabular}        \\  \hline
\textbf{Source} & \begin{tabular}[c]{@{}l@{}}应当注意的是，重音切记过多。过多则显示不了孰轻孰重。\\ 
It is worth noting that too much stress should be remembered. Too much stress can not show which is more important. \end{tabular} \\ 
\textbf{Ref. 1} & \begin{tabular}[c]{@{}l@{}}应当注意的是，重音\textcolor{red}{切忌}过多。过多则显示不了孰轻孰重。\\ 
It is worth noting that too much stress should be \textcolor{red}{avoided}. Too much stress can not show which is more important.  \end{tabular}        \\
\textbf{Ref. 2} & \begin{tabular}[c]{@{}l@{}}应当注意的是，重音切记\textcolor{red}{不要}过多。过多则显示不了孰轻孰重。\\ 
It is worth noting that \textcolor{red}{avoiding} too much stress should be remembered. Too much stress can not show which is more important.  \end{tabular}        \\
\hline \hline
\multicolumn{2}{c}{\textbf{Domain: \textsc{Thesis}}} \\ \hline
\textbf{Source} & \begin{tabular}[c]{@{}l@{}}目前应用最为广泛的词干提取方法为波特词干算法（Poter-Stemmer），它基于后缀进行玻璃。\\ At present, the most widely used stemming method is the Poter-Stemmer algorithm, which is based on the suffix for glass. \end{tabular} \\ 
\textbf{Ref. 1} & \begin{tabular}[c]{@{}l@{}}目前应用最为广泛的词干提取方法为波特词干算法（Poter-Stemmer），它基于后缀进行\textcolor{red}{剥离}。\\ At present, the most widely used stemming method is the Poter-Stemmer algorithm, which is based on the suffix for \textcolor{red}{stripping}.  \end{tabular}
\\\hline
\textbf{Source} & \begin{tabular}[c]{@{}l@{}}word2vec的基本结构是一个输入隐藏输出的三层神经网络。\\ The basic structure of word2vec is a three-layer neural network with input hidden output.\end{tabular} \\ 
\textbf{Ref. 1} & \begin{tabular}[c]{@{}l@{}}word2vec的基本结构是一个\textcolor{red}{包含输入层、隐藏层和输出层}的三层神经网络。\\ The basic structure of word2vec is a three-layer neural network \textcolor{red}{including the input layer, hidden layer and output layer}.  \end{tabular}        \\ 
\textbf{Ref. 2} & \begin{tabular}[c]{@{}l@{}}word2vec的基本结构是一个\textcolor{red}{由输入层、隐藏层和输出层组成}的三层神经网络。\\ The basic structure of word2vec is a three-layer neural network \textcolor{red}{composed of the input layer, hidden layer and output layer}.  \end{tabular}  \\ 
  \hline \hline
\multicolumn{2}{c}{\textbf{Domain: \textsc{Exam}}} \\ \hline
\textbf{Source} & \begin{tabular}[c]{@{}l@{}}止咳祛痰片，它里面的主要成分是远志、桔梗、贝母、氯化铵等配制而成的。\\ Zhike Qutan Tablet, the main components of which are mainly compounded of Milkwort, Platycodon grandiflorum, Fritillaria, \\ Ammonium chloride, etc. \end{tabular} \\ 
\textbf{Ref. 1} & \begin{tabular}[c]{@{}l@{}}止咳祛痰片，它里面的主要成分是远志、桔梗、贝母、氯化铵等\sout{配制而成的}。\\ Zhike Qutan Tablet, the main components of which are \sout{mainly compounded of} Milkwort, Platycodon grandiflorum, Fritillaria, \\ Ammonium chloride, etc.  \end{tabular}        \\ 
\textbf{Ref. 2}  & \begin{tabular}[c]{@{}l@{}}止咳祛痰片，它\sout{里面的主要成分}是远志、桔梗、贝母、氯化铵等配制而成的。\\ Zhike Qutan Tablet, \sout{the main components of} which are mainly compounded of Milkwort, Platycodon grandiflorum, Fritillaria, \\ Ammonium chloride, etc.  \end{tabular}       \\ \hline
\textbf{Source} & \begin{tabular}[c]{@{}l@{}}同学们临走时总是忘记关灯。从这一件平凡的小事中，却说明了一个大问题。\\ The students always forget to turn off the lights when they leave. From this trivial matter, shows a big problem. \end{tabular} \\ 
\textbf{Ref. 1} & \begin{tabular}[c]{@{}l@{}}同学们临走时总是忘记关灯。从这一件平凡的小事中，\textcolor{red}{我们却发现}了一个大问题。\\ The students always forget to turn off the lights when they leave. From this trivial matter, \textcolor{red}{we found} a big problem.  \end{tabular}        \\ 
\textbf{Ref. 2} & \begin{tabular}[c]{@{}l@{}}同学们临走时总是忘记关灯。\sout{从}这一件平凡的小事\sout{中，}却说明了一个大问题。\\ The students always forget to turn off the lights when they leave. \sout{From} This trivial matter shows a big problem.  \end{tabular}        \\ 
\bottomrule 
\end{tabular}
}

\caption{Annotation examples in NaSGEC. ``Source'' denotes the source sentence, ``Ref'' denotes the reference.}
\label{tab:example}
\end{table*}

\section{Annotation Examples}
\label{sec:examples}
We show some real annotation examples from NaSGEC in Table \ref{tab:example}.
We also present screenshots of all data sources of our domains in Figure \ref{fig:data:source}.




\end{CJK}
\end{document}